\documentclass[10pt,twocolumn,letterpaper]{article}

\usepackage{wacv}              

\usepackage{graphicx}
\usepackage{amsmath}
\usepackage{amssymb}
\usepackage{booktabs}

\usepackage[pagebackref,breaklinks,colorlinks]{hyperref}

\usepackage{csquotes}
\usepackage{graphics}
\usepackage{tabu}
\usepackage{multirow}
\usepackage{multicol}
\usepackage{makecell}
\usepackage{colortbl}
\usepackage{enumitem}
\usepackage[font=small]{caption}
\usepackage{ctable}

\usepackage{xcolor}

\newcommand{\concat}{\mathcal{}{||}}

\newcommand{\revise}[1]{\textcolor{black}{#1}}

\newcommand\T{\rule{0pt}{2.9ex}}       
\newcommand\B{\rule[-1.2ex]{0pt}{0pt}} 
\usepackage[export]{adjustbox}
\usepackage{adjustbox}

\usepackage[capitalize]{cleveref}
\crefname{section}{Sec.}{Secs.}
\Crefname{section}{Section}{Sections}
\Crefname{table}{Table}{Tables}
\crefname{table}{Tab.}{Tabs.}

\def\wacvPaperID{234} 
\def\confName{WACV}
\def\confYear{2025}

\begin{document}

\title{SEM-Net: Efficient Pixel Modelling for image inpainting with Spatially Enhanced SSM}

\author{
Shuang Chen$^1$ \quad Haozheng Zhang$^{1}$ \quad Amir Atapour-Abarghouei$^1$ \quad Hubert P. H. Shum$^1$$^{\dag}$ \quad \\
{\tt\small $^1$\{shuang.chen, haozheng.zhang, amir.atapour-abarghouei, hubert.shum\}@durham.ac.uk}\\
{\small $^{\dag}$Corresponding Author}
}
\maketitle

\begin{abstract}
   Image inpainting aims to repair a partially damaged image based on the information from known regions of the images. \revise{Achieving semantically plausible inpainting results is particularly challenging because it requires the reconstructed regions to exhibit similar patterns to the semanticly consistent regions}. This requires a model with a strong capacity to capture long-range dependencies. Existing models struggle in this regard due to the slow growth of receptive field for Convolutional Neural Networks (CNNs) based methods and patch-level interactions in Transformer-based methods, which are ineffective for capturing long-range dependencies.   
   Motivated by this, we propose SEM-Net, a novel visual State Space model (SSM) vision network, modelling corrupted images at the pixel level while capturing long-range dependencies (LRDs) in state space, achieving a linear computational complexity. To address the inherent lack of spatial awareness in SSM, we introduce the Snake Mamba Block (SMB) and Spatially-Enhanced Feedforward Network. These innovations enable SEM-Net to outperform state-of-the-art inpainting methods on two distinct datasets, showing significant improvements in capturing LRDs and enhancement in spatial consistency. Additionally, SEM-Net achieves state-of-the-art performance on motion deblurring, demonstrating its generalizability. Our source code is available: {\url{https://github.com/ChrisChen1023/SEM-Net}}.

\end{abstract}

\vspace{-0.7cm}
\section{Introduction}
\label{sec:intro}


Image inpainting is a highly challenging low-level vision task in computer vision due to its ill-posed nature. It aims to repair partially damaged or missing regions by leveraging information from the known areas~\cite{zhao2020uctgan}. Successful inpainting relies heavily on advanced image representation learning, particularly in capturing both short-range and long-range dependencies~\cite{deng2023context,zhao2020uctgan}, to ensure consistent reconstruction between the filled and visible contents.



Convolutional Neural Networks (CNNs) are widely used as backbone networks in image inpainting due to their strong performance in learning generalizable representations from images and their effective mining of short-range dependencies through convolution operations~\cite{yu2020region,suin2021distillation,yi2020contextual,peng2021generating}. However, their slow-grown receptive field constrains the perception of the global context and hampers the ability to capture long-range dependencies within the image. This limitation is particularly problematic for low-level vision tasks like image inpainting, where single-pixel reconstruction must preserve pixel consistency while accounting for dependencies over larger distances.
To address this limitation,
researchers have shifted towards transformer-based architectures~\cite{li2022mat,chen2021pre} to better capture the long-range dependencies (LRDs) and global structure. However, transformer-based methods suffer from quadratic computational complexity, which restricts their ability to learn spatial LRDs only at the patch level rather than the pixel level. \cite{zamir2022restormer} attempts to model images at the pixel level with transformer, but it focuses on semantic features rather than spatial relations, which means it still lacks the ability to effectively capture spatial LRDs.

\begin{figure*}[t] \centering
    \begin{tabular}{cl} 
        \rotatebox{90}{{\fontfamily{ptm}\selectfont ~~~Sample I}} &     
        \includegraphics[width=0.13\textwidth]{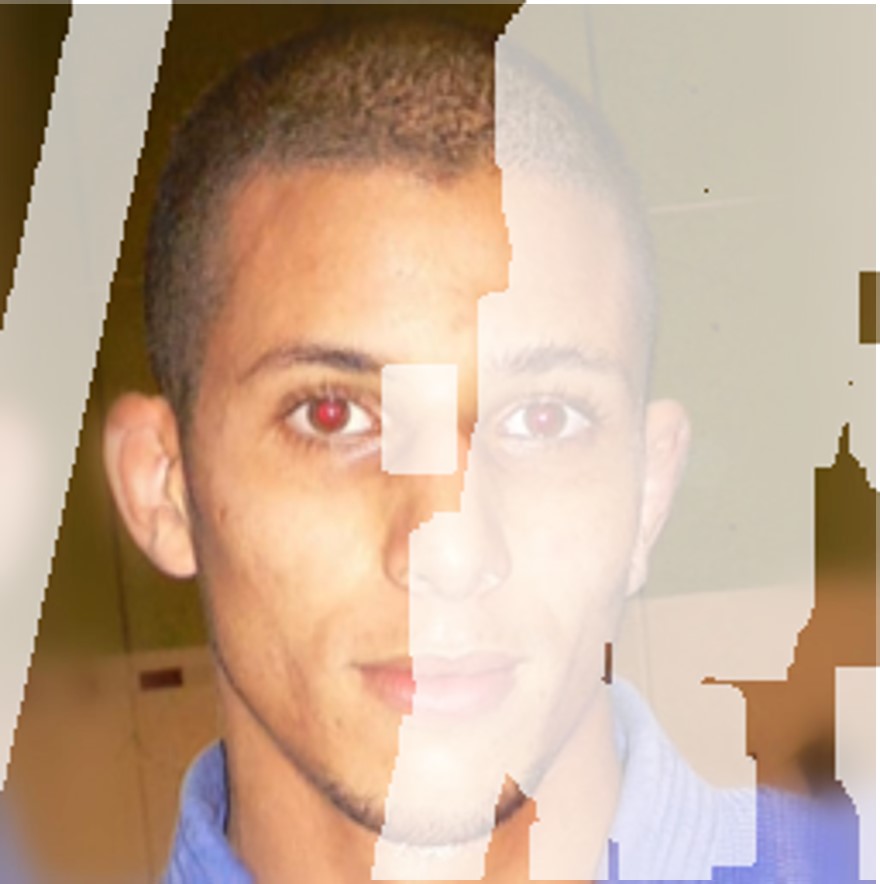}
        \includegraphics[width=0.13\textwidth]{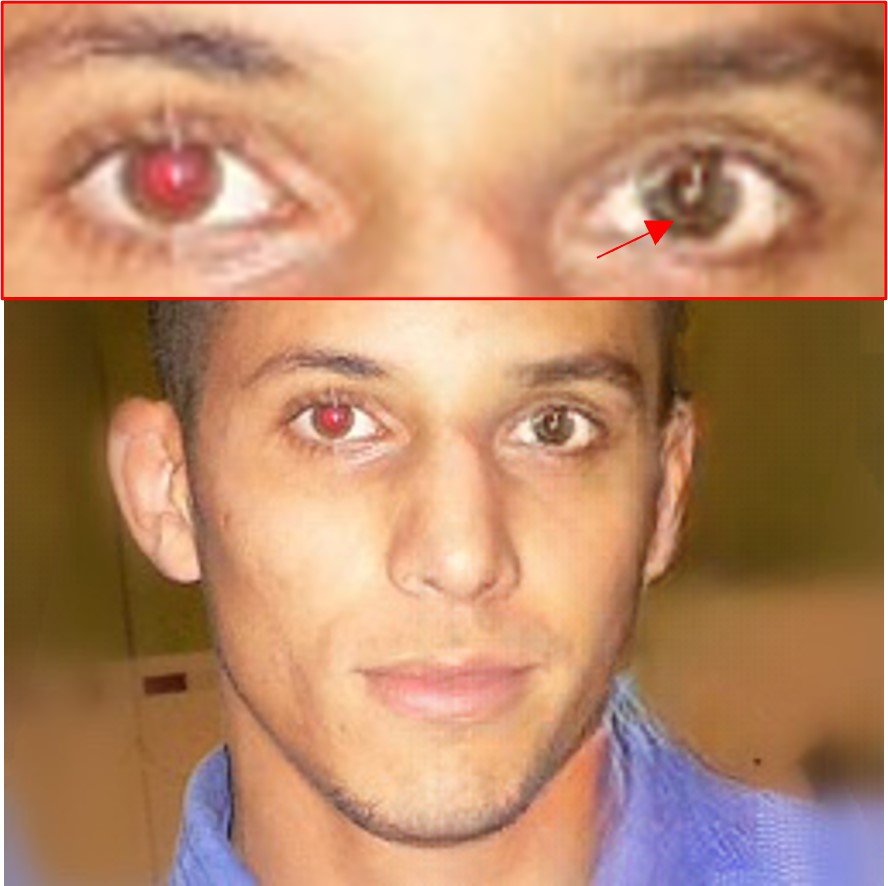}
        \includegraphics[width=0.13\textwidth]{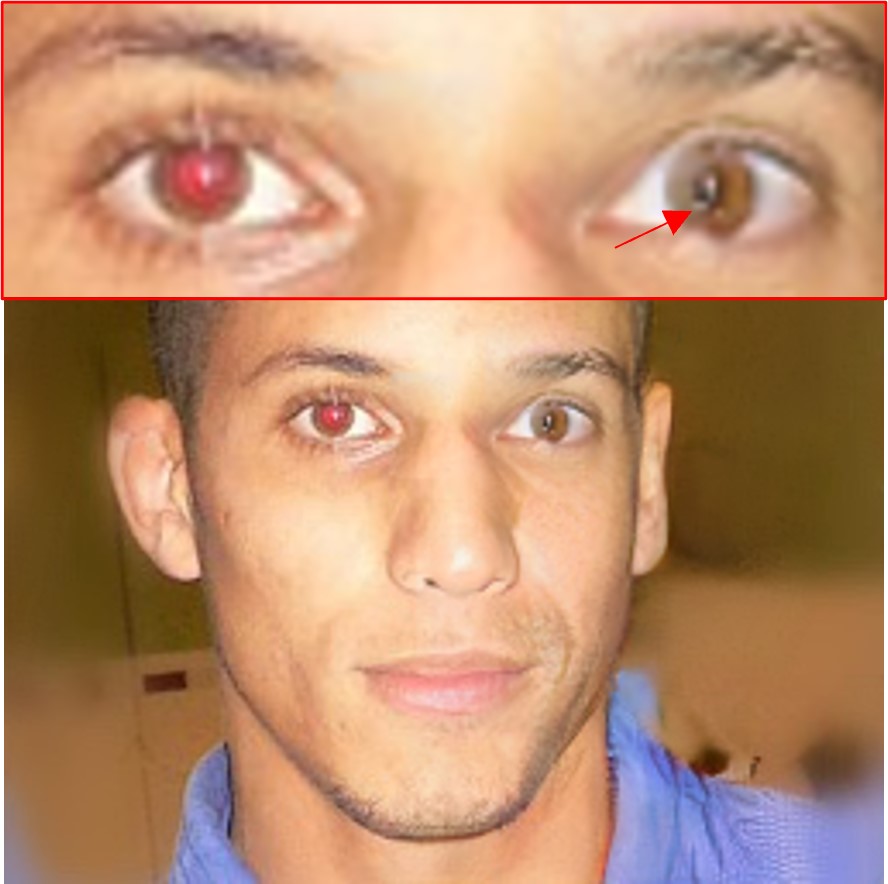}
        \includegraphics[width=0.13\textwidth]{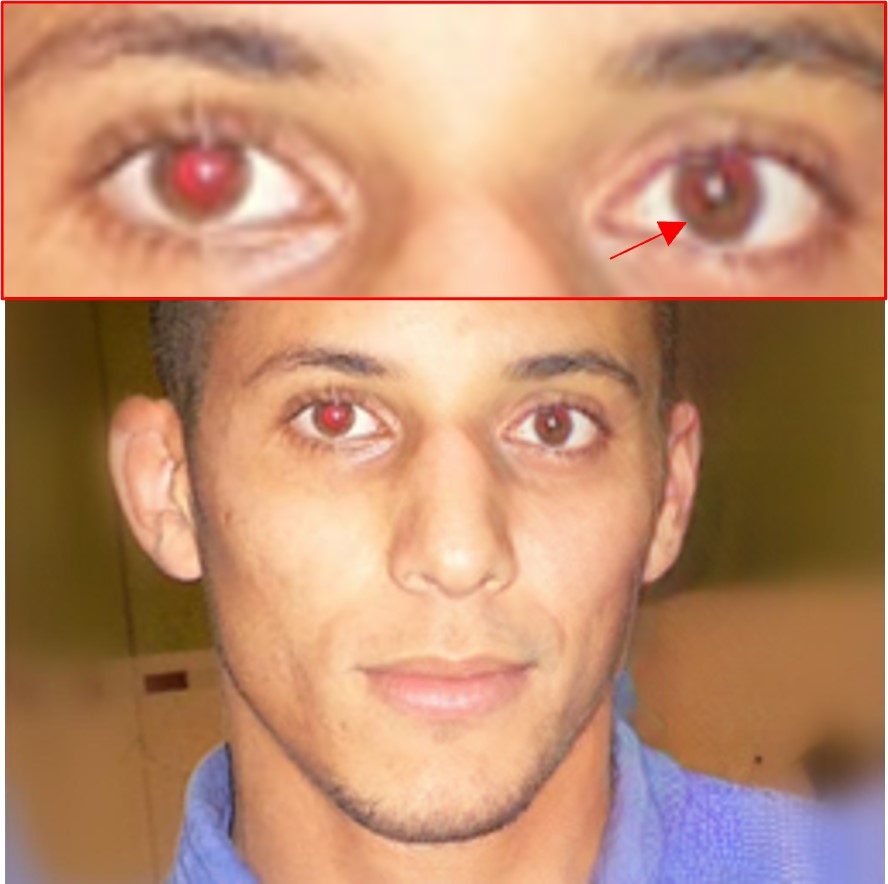}
        \includegraphics[width=0.13\textwidth]{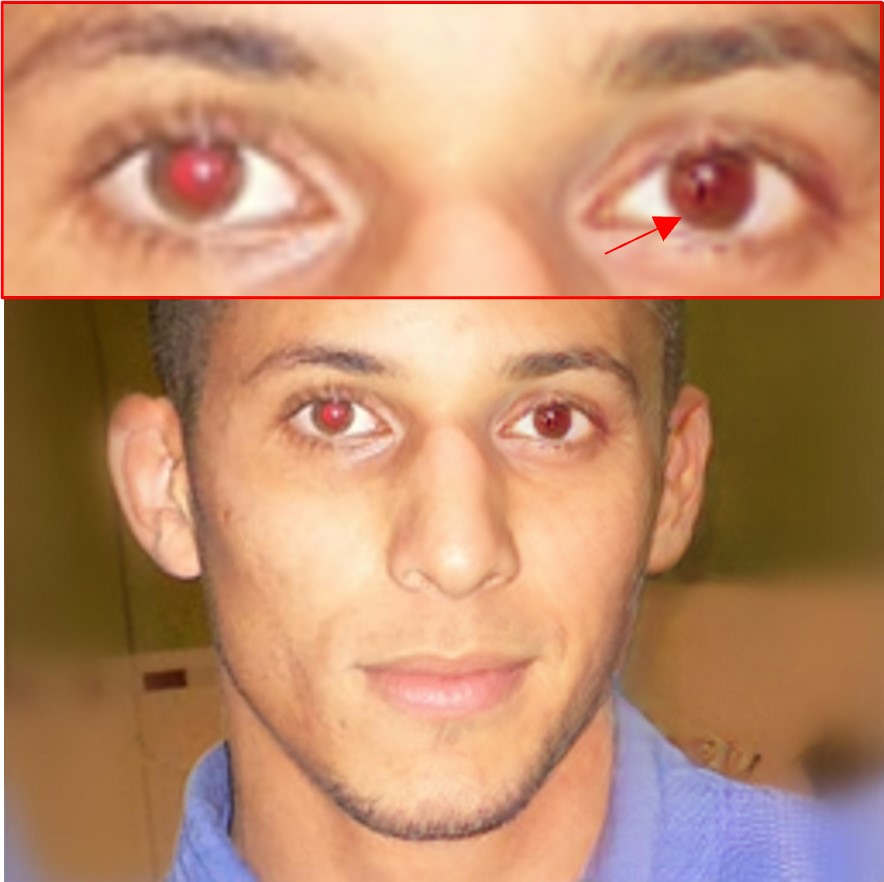}
        \includegraphics[width=0.13\textwidth]{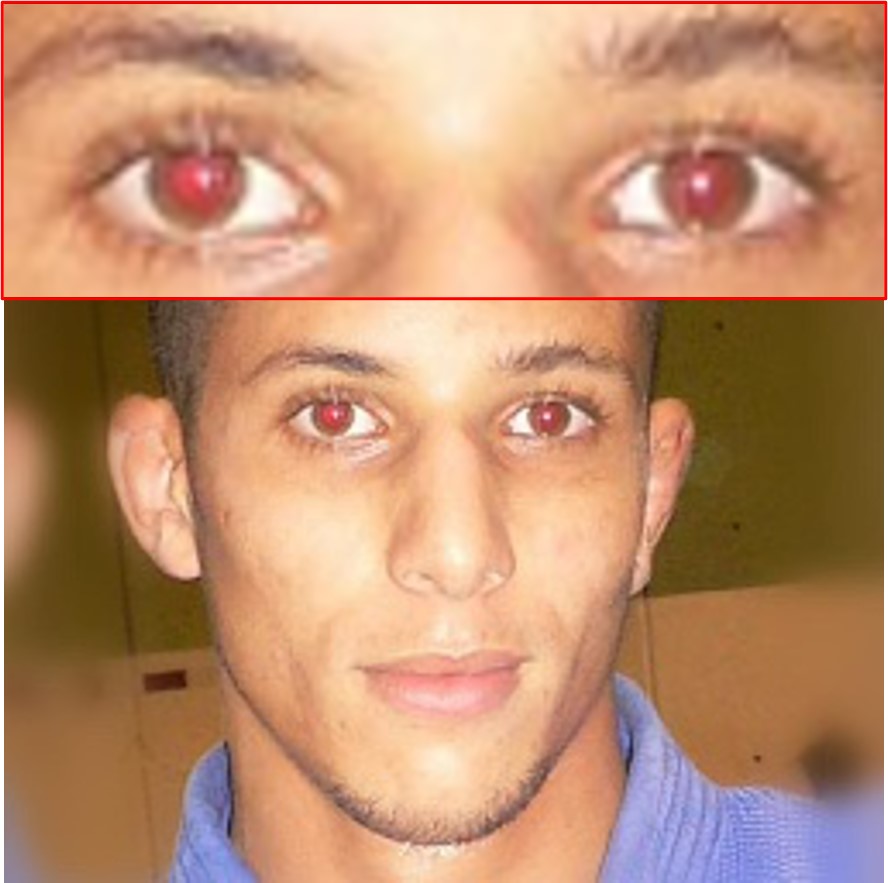} \\
    \end{tabular}
    \begin{tabular}{cl} 
        \rotatebox{90}{{\fontfamily{ptm}\selectfont ~~Sample II}} &     
        \includegraphics[width=0.13\textwidth]{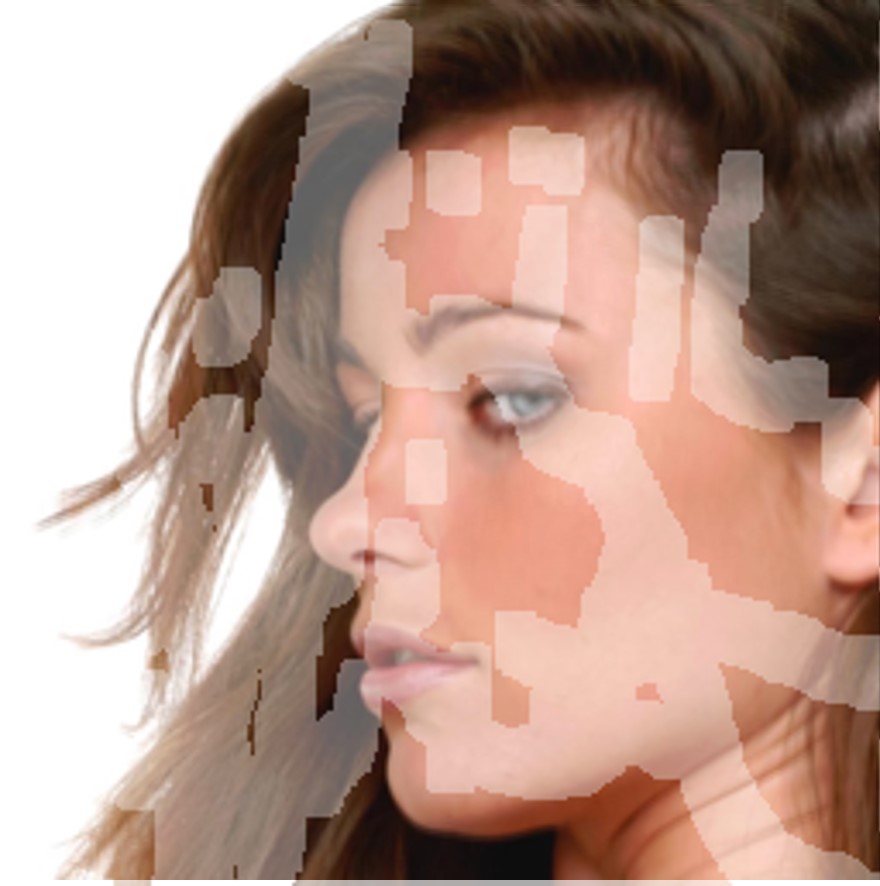}
        \includegraphics[width=0.13\textwidth]{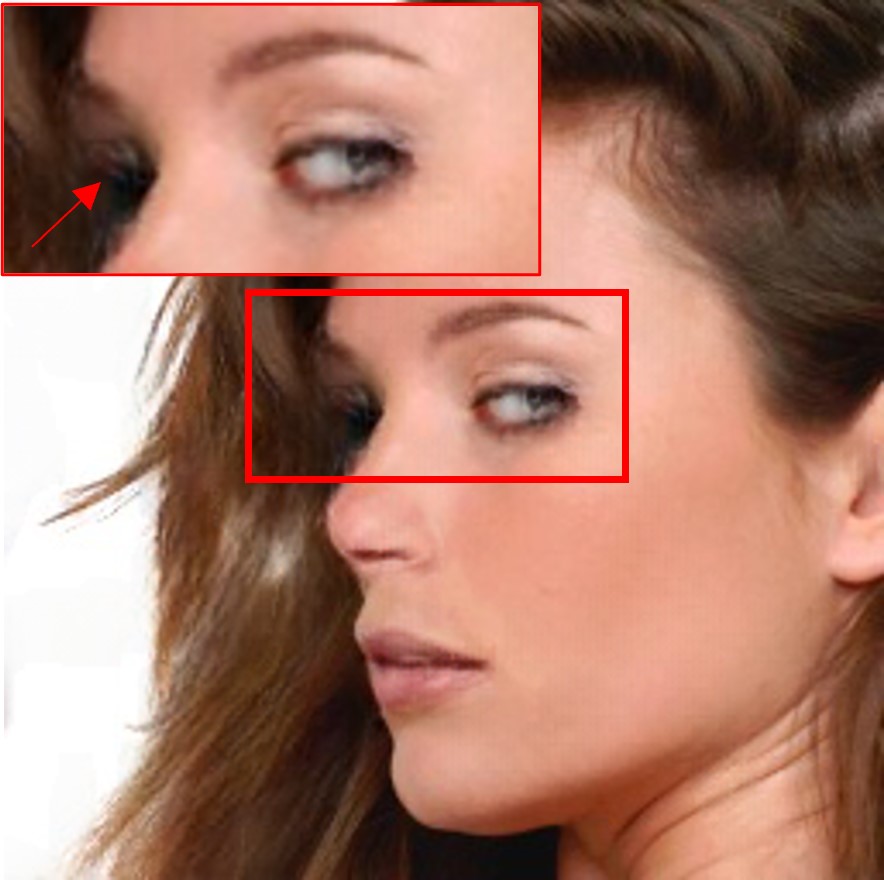}
        \includegraphics[width=0.13\textwidth]{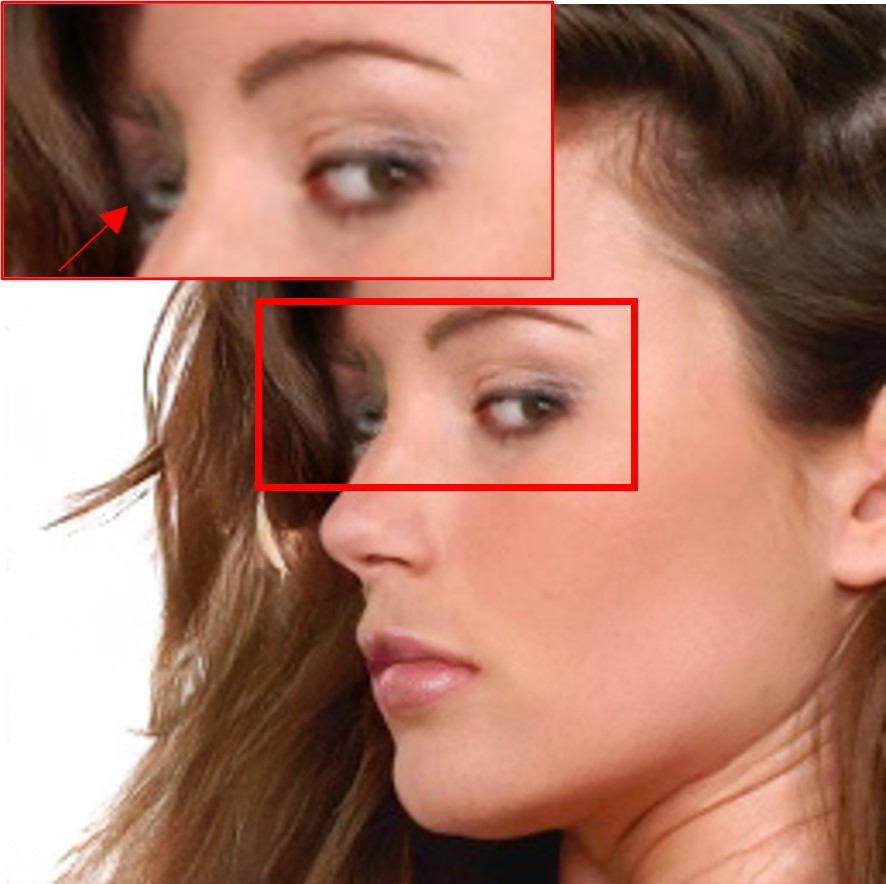}
        \includegraphics[width=0.13\textwidth]{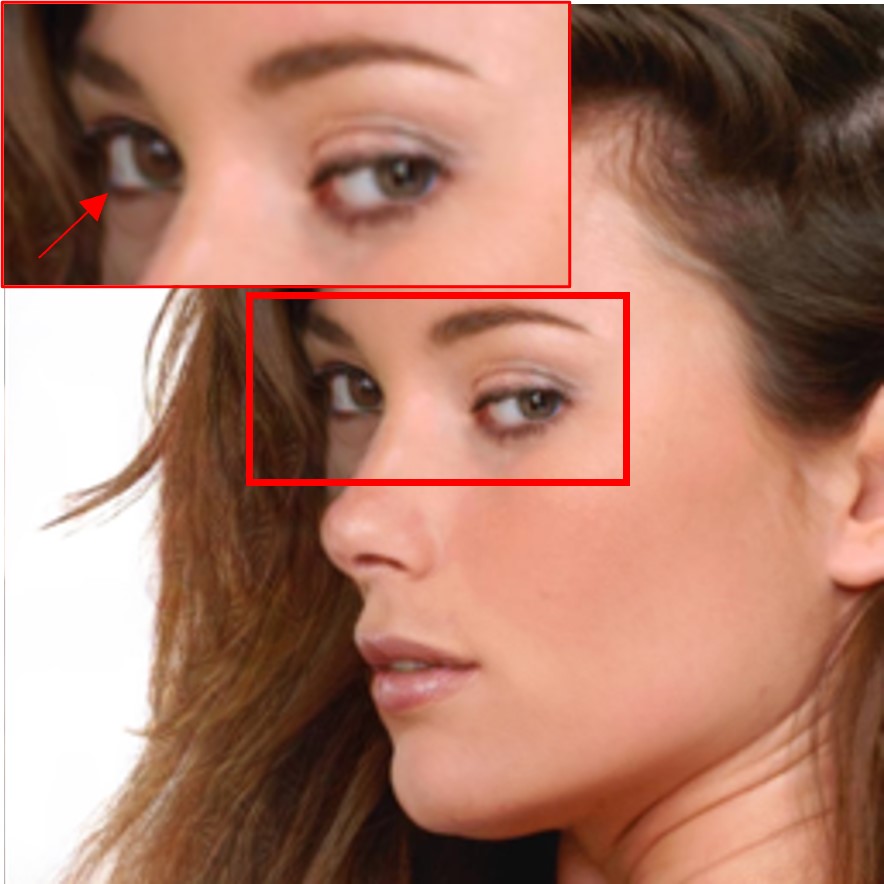}
        \includegraphics[width=0.13\textwidth]{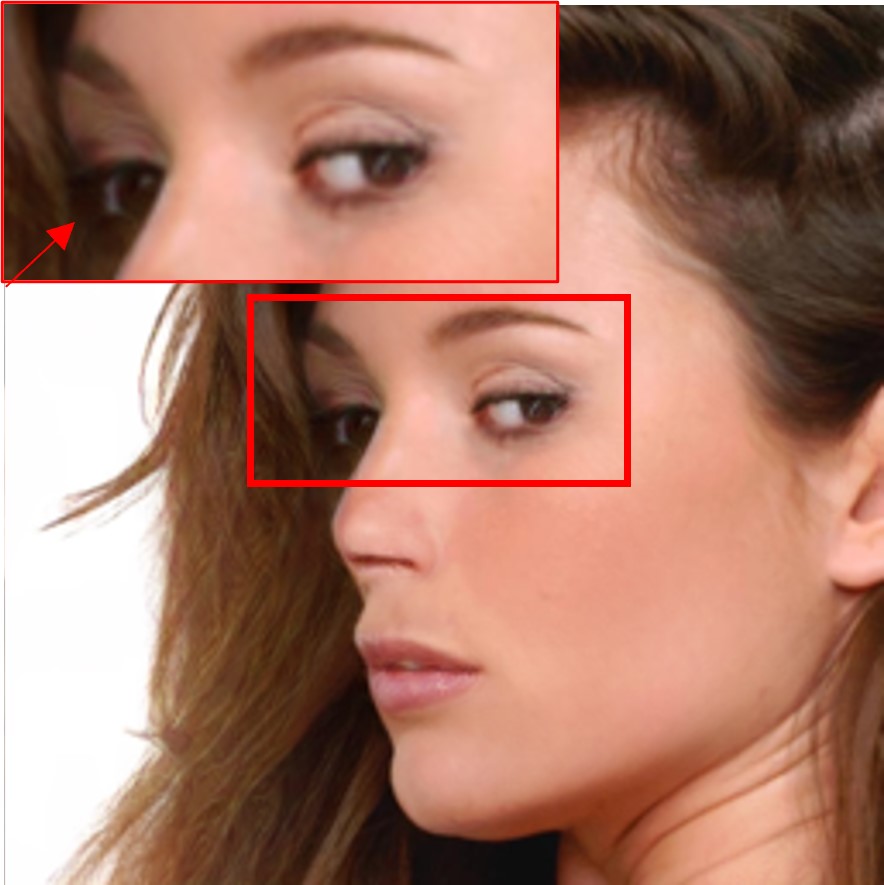}
        \includegraphics[width=0.13\textwidth]{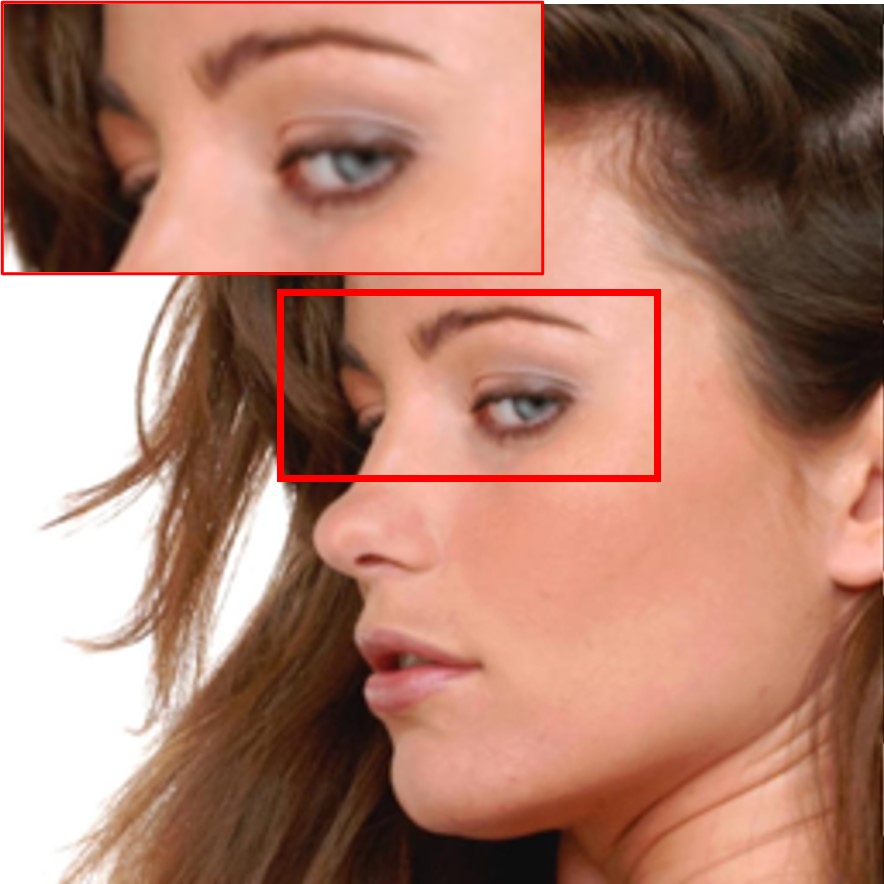}
        \\
    \end{tabular}
    \makebox[0.13\textwidth]{\footnotesize ~~~Input}
    \makebox[0.13\textwidth]{\footnotesize \text{~~~LAMA}\cite{suvorov2022resolution}}
    \makebox[0.13\textwidth]{\footnotesize ~~~MAT\cite{li2022mat}}
    \makebox[0.13\textwidth]{\footnotesize \text{~~M-Unet}}
    \makebox[0.13\textwidth]{\footnotesize \text{~~SEM-Net}}
    \makebox[0.13\textwidth]{\footnotesize ~~~GT}
    \vspace{-0.3cm}
    \caption{Comparisons with the state-of-the-art CNN-based method~\cite{suvorov2022resolution} and transformer-based method~\cite{li2022mat}. M-Unet is a variant of directly applying the Mamba model~\cite{ma2024u} followed by a feedforward network~\cite{zamir2022restormer} in a U-Net. Red boxes and arrows highlight major differences. Our SEM-Net demonstrates the strong capability to capture LRDs visualised by the consistent eye colors and patterns, and addresses the challenge of lack of spatial awareness in M-Unet. Please refer to the supplementary material for more quantitative results.}
    \vspace{-0.6cm}
    \label{fig:teaser}
\end{figure*}

\revise{LRDs are critical in image inpainting, as a lack of LRDs often results in low-quality outcomes due to insufficient context capturing.
This is evidenced by the inconsistent eye colours and patterns, as shown in the visualization of the prominent CNN-based method~\cite{suvorov2022resolution} and transformer-based method~\cite{li2022mat} (Sample I of Fig.~\ref{fig:teaser}), where the visible red eye fails to guide the accurate reconstruction of the other eye. }


One feasible solution to this challenge is from~\cite{gu2023mamba}, which proposes an emerging Selective State Space Model (SSM), known as Mamba. \revise{SSM has demonstrated its efficient and effective capacity in learning LRDs, and good adaptability in computer vision~\cite{liu2024vmamba}.}
As shown in Sample I of Fig.~\ref{fig:teaser}, directly adopting SSM~\cite{ma2024u} (M-Unet) captures LRDs effectively and achieves more consistent eye colour. 

However, as the vanilla SSM scans the data as a sequence with a single fixed direction, it lacks 2D spatial awareness, making the way to model pixels in SSM crucial. As illustrated in Sample II of Fig.~\ref{fig:teaser}, a vanilla SSM model~\cite{ma2024u} shows positional drifting of the inpainted left eye (upper than the right eye). This insight introduces two key challenges: (i) how to maintain the continuity and consistency of pixel adjacency for pixel-level dependencies learning while processing the SSM recurrence; and (ii) how to effectively integrate 2D spatial awareness to the predominant linear recurrent-based SSMs.

To solve these challenges, we propose \textbf{SEM-Net}: \textbf{S}patially-\textbf{E}nhanced SS\textbf{M} \textbf{Net}work for image inpainting, which is a simple yet effective encoder-decoder architecture incorporating four-stage Snake Mamba
Blocks (SMB). The proposed SMB is assembled by two novel modules, which holistically integrate local and global spatial awareness into the model. Specifically, we introduce the Snake Bi-Directional Modelling module (SBDM) in place of vanilla SSM. It brings the crucial spatial context into a linear recurrent system, modelling images in two directions by consistently scanning each pixel with a snake shape. \revise{Moreover, we explicitly incorporate positional embedding into the sequences via a  Position Enhancement Layer (PE layer) to strengthen the long-range positional awareness and improve the sensitivity to specific parts of the sequence(e.g., masked regions)}. We further propose Spatially-
Enhanced Feedforward Network (SEFN) to complement the local spatial dependencies. aiming to leverage spatial information stored in the feature before SBDM, to refine the feature after SBDM with a gating mechanism.

Comparative experiments show that SEM-Net outperforms state-of-the-art approaches across two distinct datasets, i.e, CelebA-HQ \cite{karras2017progressive} and Places2 \cite{zhou2017places}. Detailed qualitative comparison demonstrates that our method achieves a significant improvement in capturing spatial LRDs while preserving better spatial structure. In addition, SEM-Net achieves state-of-the-art performance on two motion-deblurring datasets, further demonstrating our method's generalizability in image representation learning.

Our main contributions are summarized as follows:
\vspace{-0.2cm}
\begin{itemize}
    \item We propose a novel U-shaped Spatially-Enhanced SSM architecture focused on capturing short- and long-range spatial dependencies in image inpainting. To the best of our knowledge, SEM-Net is the first SSM-based model in this research field. 
    \item We propose a Snake Mamba Block (SMB), involving a Snake Bi-Directional Modelling (SBDM) module and a Position Enhancement Layer (PE layer), to implicitly integrate crucial spatial context awareness into a linear recurrent SSM, and explicitly enhance the long-range positional awareness.
    \item We propose a Spatially-Enhanced Feedforward Network (SEFN) to complement local spatial dependencies learning among pixels, enhancing the spatial awareness throughout image representation learning. 
\end{itemize}
\section{Related Work}
\subsection{Image Inpainting} Image inpainting is a classic ill-posed low-level vision task that requires inferring the missing or damaged areas of the image based on the known pixels. 
The rapid advancements in CNNs have led to the introduction of diverse techniques that have significantly improved image inpainting by using CNN-based encoder-decoder architectures~\cite{yan2018shift, yu2020region,suin2021distillation} or CNN-based Generative Adversarial Networks (GANs)~\cite{xu2014deep,yu2018generative,iizuka2017globally,yi2020contextual,peng2021generating,sargsyan2023mi}. Nonetheless, limited receptive fields of convolution hinder the capturing of long-range dependencies~\cite{xu2021hip,zamir2022restormer}, which motivates the researchers to enlarge the receptive fields by employing convolution in frequency domain~\cite{suvorov2022resolution,chu2023rethinking} or developing transformer-based models~\cite{chen2024hint,li2022mat,chen2021pre,liang2021swinir}. However, practical applications are limited by the computational complexity of self-attention, specifically, as it is still challenging to achieve pixel-level self-attention at relatively high resolutions~\cite{cui2023image}. In this paper, we primarily focus on the field of selective SSM for achieving image pixel-level LRDs learning with strong spatial awareness.

\subsection{SSM in Computer Vision} SSMs serve as fundamental models in various fields, including control theory and computational neuroscience. However, the application of SSMs in deep learning is limited due to the risk of vanishing gradients when using the linear first-order Ordinary Differential Equations to solve an exponential function~\cite{gu2021combining}. To solve this problem, Gu et al.~\cite{gu2021efficiently} proposed a structured SSM model with the HiPPO framework~\cite{gu2020hippo} to address the significant computational challenges and model LRDs with rigorous theoretical proofs. Recently, \cite{gu2023mamba} further proposed a selective structured SSM, namely Mamba, to allow for context-based reasoning using long sequence inputs. Inspired by them, several researchers have begun to explore the selective SSMs in computer vision tasks, including image segmentation~\cite{ma2024u}, image classification and object detection~\cite{liu2024vmamba,zhu2024vision}. However, these works have not effectively leveraged the potential of Mamba in capturing long-range pixel-level spatial dependencies for image representation learning. \revise{\cite{hu2024zigma} uses a zigzag scanning approach to improve spatial continuity in patched images. However, the fixed scanning directions can result in a loss of joint information between pixels, particularly as the patch size increases.} 

\subsection{Preliminary}
\label{pre_ssm}

While SSMs are generally regarded as linear time-invariant systems that map a 1-dimensional sequence $x(t)\in \mathbb{R}$ to response $y(t)\in \mathbb{R}$ via an implicit latent state $h(t)\in \mathbb{R}^N$ (Eq.\ref{eq1}), the structured SSMs use zero-order hold discretisation rule (Eq.\ref{eq2}) to transform the continues parameter $(\Delta,\textbf{\textit{A}},\textbf{\textit{B}})$ to discrete parameters $(\overline{\textbf{\textit{A}}},\overline{\textbf{\textit{B}}})$ for allowing efficient linear recurrence in (Eq.\ref{eq3}).
\vspace{-0.3cm}
\begin{align}
&h'(t) = \textit{\textbf{A}}h(t) + \textbf{\textit{B}}x(t), \quad y(t) = \textbf{\textit{C}}h(t), \label{eq1}\\  
&\overline{\textit{\textbf{A}}} = \exp(\Delta \textbf{\textit{A}}), \quad \overline{\textbf{\textit{B}}} = (\Delta \textbf{\textit{A}})^{-1}(\exp(\Delta \textbf{\textit{A}}) - \textit{\textbf{I}}) \cdot \Delta \textbf{\textit{B}},\label{eq2} \\   
&h_t = \overline{\textbf{\textit{A}}}h_{t-1} + \overline{\textbf{\textit{B}}}x_t, \quad y_t = \textit{\textbf{C}}h_t, \label{eq3} 
\end{align}
where $\textbf{\textit{A}}\in \mathbb{R}^{N\times N}$, $\textbf{\textit{B}}\in \mathbb{R}^{N\times 1}$, $\textbf{\textit{C}}\in \mathbb{R}^{1\times N}$, and $\Delta$ is the step size.

To enhance the hidden state dimension while avoiding the trade-offs in speed and memory, Gu et al.~\cite{gu2023mamba} further proposes an architecture (i.e., Mamba) with a selection mechanism on SSMs to transform parameters $\Delta$,\textbf{\textit{B}},\textbf{\textit{C}}, into functions that depend on the input, such that introducing a length dimension to these parameters, making the model changed from time-invariant to time-varying.


\begin{figure*}
  \centering
  \includegraphics[height=3.7cm]{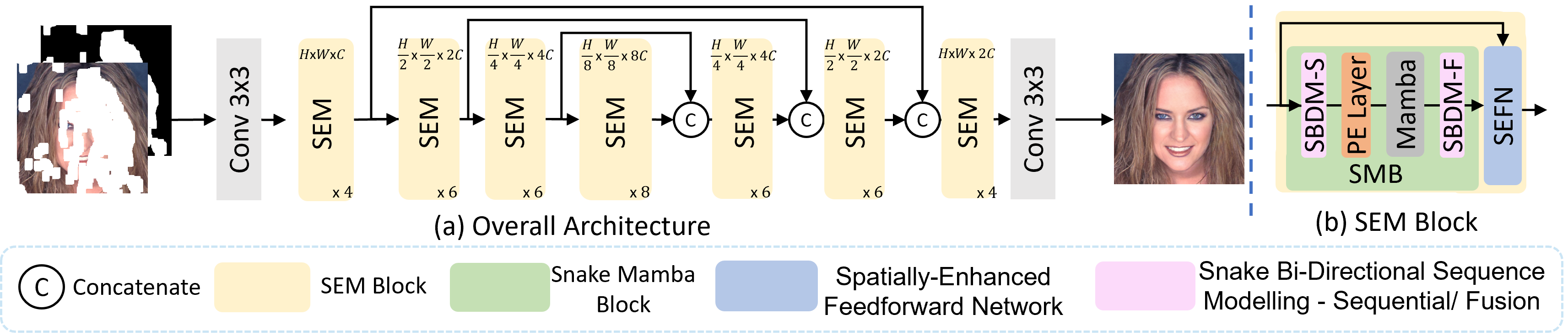}
  \vspace{-0.4cm}
  \caption{(a) Architecture overview of the proposed SEM-Net with multi-scale SEM blocks. (b) The details in each SEM block with core designs in SMB and SEFN, which holistically enhance the spatial awareness and improve the capability to capture LRDs.}
  \label{fig:overview}
\end{figure*}

\section{Spatially-Enhanced Mamba Network}
Given an image $I$ with pixels $\hat{x}_{1},\hat{x}_{2},..., \hat{x}_{N\times N}$ in $N\times N$ resolution. Image inpainting is a task for learning the mapping from the input masked image ${I}_{in}=\text{concat}[{I}\odot{M}, M]$ to the semantically accurate output image $I_{out}$, where $M$ is the mask. 
The overall pipeline of the proposed SEM-Net is illustrated in Fig.~\ref{fig:overview}. Our framework comprises two key components to address the two identified challenges in a synergistic manner. \revise{The first component, a Snake Mamba Block (Sec.~\ref{method:snake}), aims at effectively preserving the continuity and consistency of pixel adjacency for pixel-level dependency learning during the linear recurrence in SSMs.} The second component, a Spatially-Enhanced Feedforward Network (Sec.~\ref{method:sefn}), is proposed to further complement the 2D spatial awareness of the 1D linear recurrent based SSMs. 

Our SEM-Net adopts the encoder-decoder based U-Net architecture formed with four-stage SEM blocks to learn hierarchical multi-scale representation. Given a masked image ${I}_{in}\in \mathbb{R}^{H\times W\times 3}$, where $H\times W$ is spatial dimension and 3 denotes the RGB channels. SEM-Net first employs a $3\times 3$ convolution to extract low-level feature embedding $\mathbf{h_0}\in \mathbb{R}^{H\times W\times C}$. Then, these features $\mathbf{h_0}$ pass through the four-scale encoder SEM blocks, which gradually decrease in spatial size while increasing in channel capacity, to generate latent features $\mathbf{h_l}\in \mathbb{R}^{\frac{H}{8}\times \frac{W}{8}\times 8C}$. Next, the decoder takes $\mathbf{h_l}$ to progressively reconstruct high-resolution representations. Every stage contains multiple SEM blocks, each SEM block has a pair of proposed Snake Mamba Block (SMB) and Spatially-Enhanced Feedforward Network (SEFN) for refining image representation learning while effectively capturing spatial LRDs. During this process, we use skip connections to link mirrored SEM blocks at the end of each stage and use a $1\times 1$ convolution to half the channels after each connection, preserving the shared features learned by the encoder and then supporting the decoder. The cost-efficient pixel-unshuffle and pixel-shuffle operations~\cite{shi2016real} are employed to achieve feature downsampling and upsampling, respectively. In the final step, a convolutional layer projects the decoded features to the output.

\subsection{Snake Mamba Block}
In each snake mamba block (SMB), we propose a holistic framework to preserve continuity and ensure the comprehensiveness of pixel adjacency for pixel-level dependency learning during 1D linear recurrence in SSMs. This is achieved through two novel designs: the implicit Snake Bi-Directional Modelling (SBDM) and the explicit Position Enhancement Layer (PE layer).

\subsubsection{Snake Bi-Directional Modelling}
\label{method:snake}
\begin{figure*}[t]
  \centering
  \includegraphics[height=4.5cm]{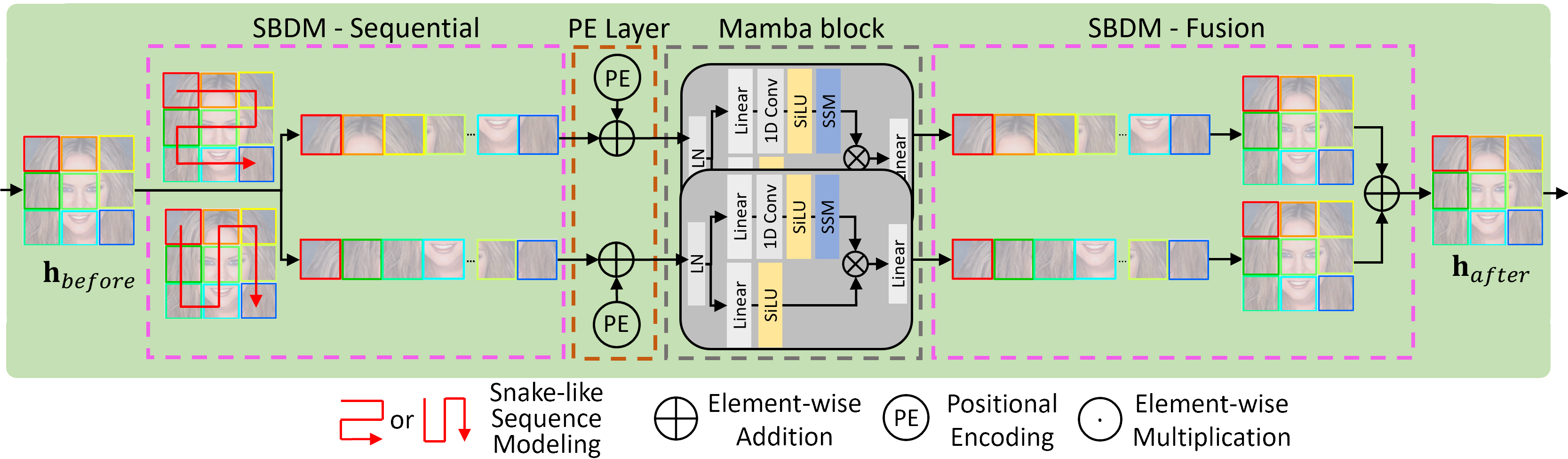}
  \vspace{-0.2cm}
  \caption{The architecture of proposed SMB. The input feature is modelled to sequences in two directions with snake-like traverses in SBDM-Sequential, enhancing the spatial awareness implicitly. Then, the PE layer explicitly enhances the long-range positional awareness through positional embeddings. The features after Mamba are restructured and aggregated by SBDM-Fusion to generate the output.}
  \vspace{-0.4cm}
  \label{fig:smb}
\end{figure*}
Directly leveraging the predominant 1D linear SSMs by feeding the flattened spatial features is prone to an inevitable loss of pixel adjacency continuity and spatial information, resulting in a degradation in image representation learning. To alleviate this challenge, SBDM mainly contains two sequence modelling techniques: snake-like sequence modelling and bi-directional sequence modelling. 

\noindent \textbf{Snake-like Sequence Modelling}. 
Snake-like sequence modelling aims to maintain the continuity in pixel adjacency when flattening spatial features across each channel from a shape of $H\times W$ to $1\times HW$. This is crucial as we observe that the conventional flattening operation directly connects the end of one row to the start of the next, forcing SSMs to recognize recurrent connections between spatially distant pixels rather than adjacent ones, leading to a loss of pixel adjacency continuity and constrains the dependency-reasoning capacity. To address this issue, our snake-like sequence modelling ensures consistent connections among neighboring pixels both within and across rows by reordering pixels and concatenating rows, illustrated by the red arrows in Fig.~\ref{fig:smb}.

Specifically, given an input feature ${{\mathbf{h}}_{in} \in \mathbb{R}^{{H}\times {W}\times {C}}}$, where $H$ is the number of rows (lines), $W$ is the number of columns (pixels in a line), and $C$ is the dimension for each pixel. ${p}_{i,j} \in \mathbb{R}^{{1}\times {1}\times {C}}$ denotes the pixel value at the position of $i$-th row and $j$-th column. Then, the horizontal snake-like sequence modelling process is represented as:
\begin{equation}
S_{i}=\left\{\begin{aligned}
& \left[p_{i, 0}, p_{i, 1}, \ldots, p_{i, W-1}\right], & i=0,2,4, \ldots, \\
& \left[p_{i, W-1}, p_{i, W-2}, \ldots, p_{i, 0}\right], & i=1,3,5, \ldots,
\end{aligned}\right.
\label{equ:snake}
\end{equation}
\begin{equation}
\begin{aligned}
S = \text{concat}[{S}_{0}, {S}_{1}, {S}_{2}, {S}_{3}, \dots ,{S}_{H-1}],
\end{aligned}
\label{equ:full_seq}
\end{equation}
where the 1D sequence \textit{S} maintains the pixel adjacency continuity by concatenating the sequences $S_i$ for $i\in [0,H-1]$, each $S_i$ represents the reordered pixel position in that row.

\vspace{0.15cm}
\noindent \textbf{Bi-directional Sequence Modelling} 
To further complement the comprehensiveness of pixel adjacency and implicitly enhance spatial awareness, we propose a bi-directional sequence modelling involving two processes: SBDM-Sequential (SBDM-S) and SBDM-Fusion (SBDM-F). As shown in~\ref{fig:smb}, 
SBDM-S simultaneously traverse pixels in a snake-like manner in two directions: horizontally and vertically across all pixels, enabling the SMB to generate sequences that capture discriminative dependencies. Specifically, 
in a snake-like manner, SBDM-S vertically traverses pixels to 1-D sequences  $S = \text{concat}[{S}_{0}, {S}_{1}, \dots ,{S}_{H-1}]$, and horizontally traverses pixels to 1-D sequences $T = \text{concat}[{S}_{0}^\top, {S}_{1}^\top, \dots ,{S}_{W-1}^\top]$, where each ${S}_{j}^\top$ for $j\in [0,W-1]$ contains reordered pixels in that column.
These two directions are designed since they are spatially complementary to each other and are computationally efficient in multi-directional traversals. After processing through Mamba, SBDM-F restructures the 1D sequences back to 2D via the inverse function of Eq.~\ref{equ:full_seq} and fuse them by element-wise aggregation to retain their spatial information, enriching the spatial awareness in image representation learning. 

\vspace{0.15cm}
\noindent \textbf{Position Enhancement Layer}
To further explicitly complement the implicit approach of SBDM in enhancing spatial dependency reasoning, we propose a simple yet effective strategy of integrating 1D positional embeddings to enhance position awareness.
Specifically, we incorporate the 1D positional embeddings directly into 1D sequences in the position enhancement layer (PE layer) before processing with Mamba, assigning absolute positional information to each element within the sequences for providing the positional context and maintaining the pixel adjacency relationships. Formally, assume $S(n)$ for $n \in [0, N^2-1] \cap \mathbb{Z}$ is the element at positional coordinate $n$, $PE(n)$ is the corresponding cosine positional embedding~\cite{vaswani2017attention}. Then, the elements in 1D sequence $S$ with 1D positional embeddings ${\bar{S}}{(n)}$ are integrated by aggregation:
\begin{equation}
    {\bar{S}}{(n)} = {S}{(n)} + {PE}{(n)}, n=0,1,2,3, \ldots, N^2-1.
\end{equation}

\begin{figure*}[h]
  \centering
  \includegraphics[height=4.0cm]{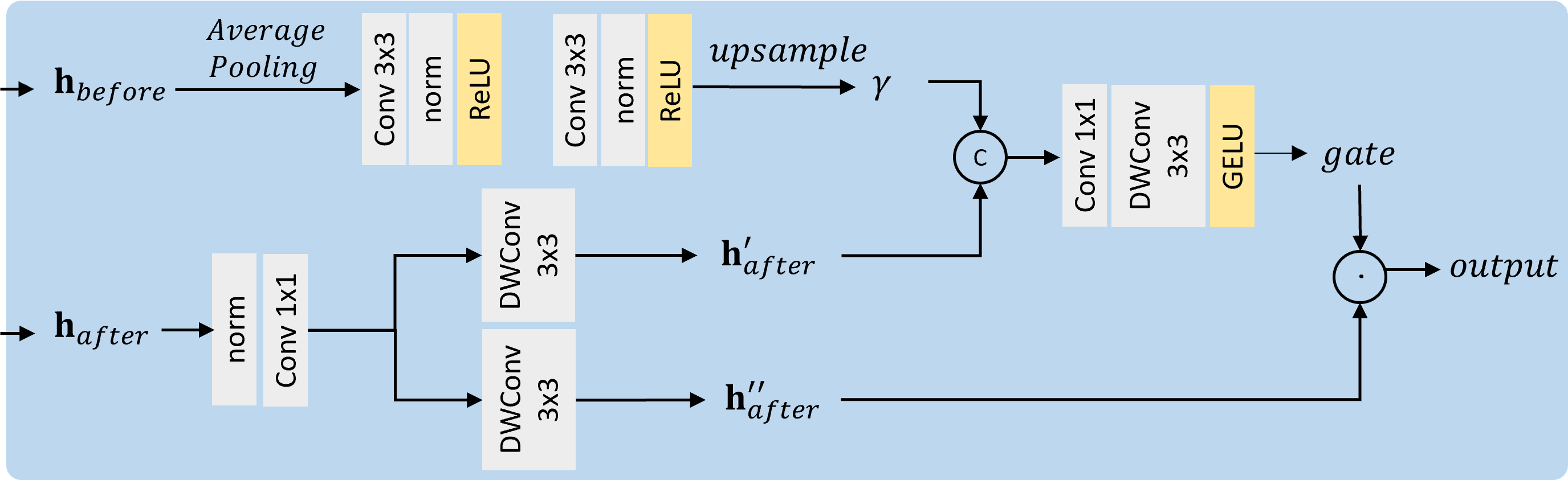}
  \vspace{-0.2cm}
  \caption{The architecture of proposed Spatially-Enhanced Feedforward Network (SEFN)}
  \label{fig:sefn}
\end{figure*}

\subsection{Spatially-Enhanced Feedforward Network}
\label{method:sefn}
To complement local spatial information in regions spanning multiple rows and columns that are subject to inherent design limitations of SSMs, we propose a Spatially-Enhanced Feedforward Network (SEFN) for refining spatial awareness in image representation learning. The key idea of SEFN lies in leveraging spatial information extracted from the feature representations prior to the SEM block, subsequently applying it in a gating mechanism to inform the features post-SMB, thereby facilitating the integration of spatial awareness and LRDs learning to the entire SEM block.


\revise{Specifically, SEFN first snatches ${\mathbf{h}}_{before}$ and ${\mathbf{h}}_{after}$ at the entrance and exit of the Mamba block. Then, SEFN uses the average pooling to expand the receptive field, followed by two \{$Conv\text{-}LN\text{-}ReLU$\} blocks to capture a broader spatial perception. The subsequent upsampling yields a spatial awareness indicator $\gamma$ preserving spatial relationships from ${\mathbf{h}}_{before}$.  The gating mechanism starts from ${\mathbf{h}}_{after}$, which is divided into ${\mathbf{h}}_{after}'$ and ${\mathbf{h}}_{after}''$. The ${\mathbf{h}}_{after}'$ is informed by $\gamma$ to form a `$gate$' via a linear transformation and a GELU non-linear activation. `$gate$' then modulates ${\mathbf{h}}_{after}''$ through a point-wise product, significantly enhancing the spatial awareness of ${\mathbf{h}}_{after}''$. The whole process is formulated as:}
\begin{align}
&{{\mathbf{h}}_{after}'} = {W_{d3}}{W_1}{LN({\mathbf{h}}_{after})},\\
&{{\mathbf{h}}_{after}''} = {W'_{d3}}{W'_1}{LN({\mathbf{h}}_{after})},\\ 
& \gamma = {Up}({f({AveragePooling}({\mathbf{h}}_{before}))}),\\
&{gate} = {GELU}({W_{d3}}{W_{1}}{\gamma}\concat{{\mathbf{h}}_{after}'}),\\
& output = {gate} \odot {{\mathbf{h}}_{after}''},
\end{align}
\revise{where ${W_1}$, ${W'_1}$ are $1\times1$ convolutions, ${W_{d3}}$, ${W'_{d3}}$ are $3\times3$ depth-wise convolutions to reduce computational cost while refining features, $LN$ is a layer normalization, $f$ denotes two \{$Conv$-$LN$-$ReLU$\} blocks, ${Up}$ is upsampling.}

\section{Experiments}
We quantitatively and qualitatively prove the superiority of our proposed inpainting method by comparing it with the state-of-the-art methods on two widely used datasets~\cite{zheng2022bridging,wan2021high,li2022mat}: CelebA-HQ~\cite{karras2017progressive} and Places2-Standard~\cite{zhou2017places} in Sec.~\ref{exp:quati} and~\ref{exp:quali}. We carefully evaluate each of the proposed novelties by a comprehensive ablation and component analysis in Sec.~\ref{exp:abla}. In Sec.~\ref{exp:gen}, we demonstrate the capability of our model in generalising to both higher resolution and unseen images. In addition, we further evaluate the image representation learning capability and generalisation ability of SEM-Net by directly applying it to another low-level vision task - image motion debluring. The implementation setting is detailed in the supplementary material.

\vspace{0.15cm}


\vspace{0.15cm}
\noindent \textbf{Baselines and Metrics}
We choose the following baselines for inpainting comparison: CNN-based methods with DeepFill v1~\cite{yu2018generative}, DeepFill v2~\cite{yu2019free}, CTSDG~\cite{guo2021image} and \textit{MISF}~\cite{li2022misf}, WaveFill~\cite{yu2021wavefill} and \textit{LAMA}~\cite{suvorov2022resolution}; Transformer-based methods with MAT~\cite{li2022mat} and \textit{CMT}~\cite{ko2023continuously}; Expensive diffusion models~\cite{Rombach_2022_CVPR, Lugmayr_2022_CVPR}. \textit{Italic} denotes the SOTA methods.
We followed~\cite{li2022misf, li2022mat} to evaluate our SEM-Net on Peak Signal-to-Noise Ratio (PSNR), Structural Similarity (SSIM), L1, Fréchet Inception Distance (FID)~\cite{heusel2017gans} and Perceptual Similarity (LPIPS).

\newcommand{\PSNRT}{PSNR}
\newcommand{\SSIMT}{SSIM}
\newcommand{\LT}{L1}
\newcommand{\LPIPST}{LPIPS}
\newcommand{\FID}{FID}
\newcommand{\LowExposure}{0.01\%-20\%}
\newcommand{\MiddleExposure}{20\%-40\%}
\newcommand{\HighExposure}{40\%-60\%}
\newcommand{\AllExposure}{Default}

\newcommand{\Frst}[1]{{\textbf{#1}}}
\newcommand{\Scnd}[1]{{\underline{#1}}}

\begin{table*}[h]
\centering
\captionsetup[table]{skip=4pt}
\caption{Quantitative comparison with the state-of-the-arts on CelebA-HQ (top), and Places2 (bottom). \textbf{Bold} and \underline{underline} are the best and the second-best respectively. \revise{Number of parameters (Param.) and inference time (Inf.) are based on the inpainting evaluation conducted on 256 × 256 images. $\revise{^C}$, $^T$ and $^D$ indicate CNN-based, Transformer-based and Diffusion-based methods, respectively.}}
\vspace{-0.3cm}
\resizebox{\textwidth}{!}{
    \begin{tabular}{@{\extracolsep{3pt}} c c| c c c c c| c c c c c| c c c c c@{}}
    \hline\hline
    \textbf{CelebA-HQ} & \multicolumn{1}{c}{\revise{Param.$\times10^6$}}
    & \multicolumn{5}{c}{\LowExposure} & \multicolumn{5}{c}{\MiddleExposure} & \multicolumn{5}{c}{\HighExposure} \T\\
    Method & \revise{/ Inf. Time} &\PSNRT$\uparrow$ & \SSIMT$\uparrow$ & \LT$\downarrow$ & \FID$\downarrow$ & \LPIPST$\downarrow$ & \PSNRT$\uparrow$ & \SSIMT$\uparrow$ & \LT$\downarrow$ & \FID$\downarrow$ & \LPIPST$\downarrow$ & \PSNRT$\uparrow$ & \SSIMT$\uparrow$ & \LT$\downarrow$ & \FID$\downarrow$ & \LPIPST$\downarrow$  \\
    \hline
    DeepFill v1 \cite{yu2018generative}$\revise{^C}$ & 3 / 7$ms$ & 34.2507 & 0.9047 & 1.7433 & 2.2141 & 0.1184 & 26.8796 & 0.8271 & 2.3117 & 9.4047 & 0.1329 & 21.4721 & 0.7492 & 4.6285 & 15.4731 & 0.2521\T\\

    DeepFill v2 \cite{yu2019free}$\revise{^C}$ & 4 / 10$ms$ & 34.4735 & 0.9533 & 0.5211 & 1.4374 & 0.0429 & 27.3298 & 0.8657 & 1.7687 & 5.5498 & 0.1064 & 22.6937 & 0.7962 & 3.2721 & 8.8673 & 0.1739\\

    WaveFill \cite{yu2021wavefill}$\revise{^C}$ & 49 / 70$ms$ & 31.4695 & 0.9290 & 1.3228 & 6.0638 & 0.0802 & 27.1073 & 0.8668 & 2.1159 & 8.3804 & 0.1231 & 23.3569 & 0.7817 & 3.5617 & 13.0849 & 0.1917\\
    RePaint~\cite{Lugmayr_2022_CVPR}$^D$ &552 / ~250000$ms$ & -&-& -& -& -& -& -& -& -& -& 21.8321 & 0.7791 & 3.9427 & 8.9637 & 0.1943 \\
    LaMa \cite{suvorov2022resolution}$\revise{^C}$& 51 / 25$ms$ & {35.5656} & 0.9685 & 0.4029 & 1.4309 & 0.0319 & {28.0348} & 0.8983 & {1.3722} & 4.4295 & 0.0903 & \Scnd{23.9419} & {0.8003} & {2.8646} & 8.4538 & 0.1620\\


    MISF \cite{li2022misf}$\revise{^C}$ & 26 / 10 $ms$ & 35.3591 & 0.9647 & 0.4957 & 1.2759 & 0.0287 & 27.4529 & 0.8899 & 2.0118 & 4.7299 & 0.1176 & 23.4476 & 0.7970 & 3.4167 & 8.1877 & 0.1868\\
    
    MAT \cite{li2022mat}$^T$  & 62 / 70$ms$ & 35.5466 & {0.9689} & {0.3961} & {1.2428} & {0.0268} & 27.6684 & {0.8957} & 1.3852 & {3.4677} & {0.0832} & 23.3371 & 0.7964 & 2.9816 & {5.7284} & {0.1575}\\

    CMT \cite{ko2023continuously}$\revise{^C}$ & 143 / 60$ms$ &\Scnd{36.0336} & \Scnd{0.9749} & \Scnd{0.3739} & \Scnd{1.1171} & \Scnd{0.0261} & \Scnd{28.1589} & \Scnd{0.9109} & \Scnd{1.2938} & \Scnd{3.3915} & \Scnd{0.0817} & 23.8183 & \Scnd{0.8141} & \Scnd{2.8025} & \Scnd{5.6382} & \Scnd{0.1567}\B\\
    \hline
    Ours & {163} / {240$ms$}& \Frst{36.8079} & \Frst{0.9774} & \Frst{0.3547} & \Frst{1.1070} & \Frst{0.0226} & \Frst{28.8776} & \Frst{0.9192} & \Frst{1.2198} &\Frst{3.3878} & \Frst{0.0716} & \Frst{24.4805} & \Frst{0.8240} & \Frst{2.6389} & \Frst{5.5972} & \Frst{0.1368} \B\T\\
    \hline\hline


    \textbf{Places2} & \multicolumn{1}{c}{\revise{Param.$\times10^6$}}& \multicolumn{5}{c}{\LowExposure} & \multicolumn{5}{c}{\MiddleExposure} & \multicolumn{5}{c}{\HighExposure} \T\\
     Method & \multicolumn{1}{c}{\revise{/ Inf. Time}}& \PSNRT$\uparrow$ & \SSIMT$\uparrow$ & \LT$\downarrow$ &\FID$\downarrow$ & \LPIPST$\downarrow$ & \PSNRT$\uparrow$ & \SSIMT$\uparrow$ & \LT$\downarrow$ &\FID$\downarrow$ & \LPIPST$\downarrow$ & \PSNRT$\uparrow$ & \SSIMT$\uparrow$ & \LT$\downarrow$ &\FID$\downarrow$ & \LPIPST$\downarrow$ \\
    \hline

    DeepFill v1\cite{yu2018generative}$\revise{^C}$ & 3 / 7$ms$& 30.2958 & 0.9532 & 0.6953 & 26.3275 & 0.0497 & 24.2983 & 0.8426 & 2.4927 & 31.4296 & 0.1472 & 19.3751 & 0.6473 & 5.2092 & 46.4936 & 0.3145\T\\

    DeepFill v2\cite{yu2019free}$\revise{^C}$ & 4 / 10$ms$ & 31.4725 & 0.9558 & 0.6632 & 23.6854 & 0.0446 & 24.7247 & 0.8572 & 2.2453 & 27.3259 & 0.1362 & 19.7563 & 0.6742 & 4.9284 & 36.5458 & 0.2891\\

    CTSDG \cite{guo2021image}$\revise{^C}$  &52 / 20$ms$ & 32.1110 & 0.9565 & 0.6216 & 24.9852 & 0.0458 & 24.6502 & 0.8536 & 2.1210 &  29.2158 & 0.1429 & 20.2962 & 0.7012 & 4.6870 & 37.4251 & 0.2712\\

    WaveFill \cite{yu2021wavefill}$\revise{^C}$& 49 / 70$ms$ & 29.8598 & 0.9468 & 0.9008 &30.4259 & 0.0519 & 23.9875 & 0.8395 & 2.5329 & 39.8519 & 0.1365 & 18.4017 & 0.6130 & 7.1015 & 56.7527 & 0.3395\\
    LDM~\cite{Rombach_2022_CVPR}$^D$ & 387 / ~6000 $ms$ & -&-& -& -& -& -& -& -& -& -& 19.6476 & 0.7052 & 4.6895 & 27.3619 & 0.2675\\
    Stable Diffusion$^D$$^*$ & 860 / 880 $ms$& -&-& -& -& -& -& -& -& -& -& 19.4812 & 0.7185 & 4.5729 & 27.8830 & 0.2416\\

    MISF~ \cite{li2022misf}$\revise{^C}$ & 26 / 10$ms$ & \Scnd{32.9873} & \Scnd{0.9615} & \Scnd{0.5931} & 21.7526 & 0.0357 & \Scnd{25.3843} & \Scnd{0.8681} & \Scnd{1.9460} & 30.5499  & 0.1183 & \Scnd{20.7260} & 0.7187 & 4.4383 & 44.4778 & 0.2278 \\

    LaMa \cite{suvorov2022resolution}$\revise{^C}$ & 51 / 25$ms$  & 32.4660 & 0.9584 & 0.5969 & \Scnd{14.7288} & \Scnd{0.0354} & 25.0921 & 0.8635 & 2.0048 & \Scnd{22.9381} & \Scnd{0.1079} & 20.6796 & \Scnd{0.7245} & \Scnd{4.4060} & \Scnd{25.9436} & \Scnd{0.2124}\\   

    CMT \cite{ko2023continuously}$^T$ & 143 / 60$ms$& 32.5765 & 0.9624 & 0.5915 & 22.1841 & 0.0364 & 24.9765 & 0.8666 & 2.0277 & 32.0184 & 0.1184 &  20.4888 & 0.7111 &  4.5484& 35.1688 & 0.2378 \B\\

    \hline
    Ours & {163} / {240$ms$}& \textbf{33.0106} & \textbf{0.9631} & \textbf{0.5902} & \textbf{14.5163} & \textbf{0.0328} & \textbf{25.4159} & \textbf{0.8736} & \textbf{1.9275} & \textbf{22.7814} & \textbf{0.1054} &  \textbf{20.8265} & \textbf{0.7279} &  \textbf{4.3614} & \textbf{25.7049} & \textbf{0.2120} \T\B\\
    \hline\hline
    \multicolumn{16}{l}{\large $^*$: The officially released Stable Diffusion inpainting model pretrained on high-quality LAION-Aesthetics V2 5+ dataset.}\T\\ 
    \end{tabular}
}
\label{tab:long}
\vspace{-0.5cm}
\end{table*}

\subsection{Quantitative Comparison}
\label{exp:quati}

We employ officially released models and test with the same masks~\cite{liu2018image} for fair comparisons. 
The quantitative results are shown in Tab~\ref{tab:long}. Our method consistently outperforms the state-of-the-art approaches~\cite{ko2023continuously,suvorov2022resolution,li2022misf} across all mask ratios in both CelebA-HQ and Places2 datasets. Particularly on the CelebA-HQ dataset, SEM-Net achieves (i) a substantial gain of 0.7743  (2.15\%$\uparrow$), 0.7187  (2.55\%$\uparrow$), and 0.5386  (2.25\%$\uparrow$) PSNR; (ii) and a significant reduction of 0.0192  (5.14\%$\downarrow$), 0.074  (5.72\%$\downarrow$), and 0.1636  (5.84\%$\downarrow$) L1, over the second methods~\cite{ko2023continuously,suvorov2022resolution} on three mask ratios, respectively. The improvements in these two specific metrics indicate a significant boost in the pixel-wise reconstruction accuracy. 
In addition, the LPIPS of SEM-Net appreciably drops than the second-best method~\cite{ko2023continuously} in CelebA-HQ dataset by 0.0035  (13.41\%$\downarrow$), 0.0101  (12.36\%$\downarrow$), and 0.0199  (12.70\%$\downarrow$) on three mask ratios, respectively. It demonstrates a significant improvement in high-quality image inpainting with lower perceptual differences.

\begin{figure*}[t] \centering
    \includegraphics[width=0.1\textwidth]{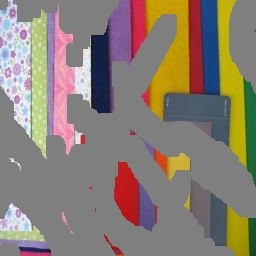}
    \includegraphics[width=0.1\textwidth]{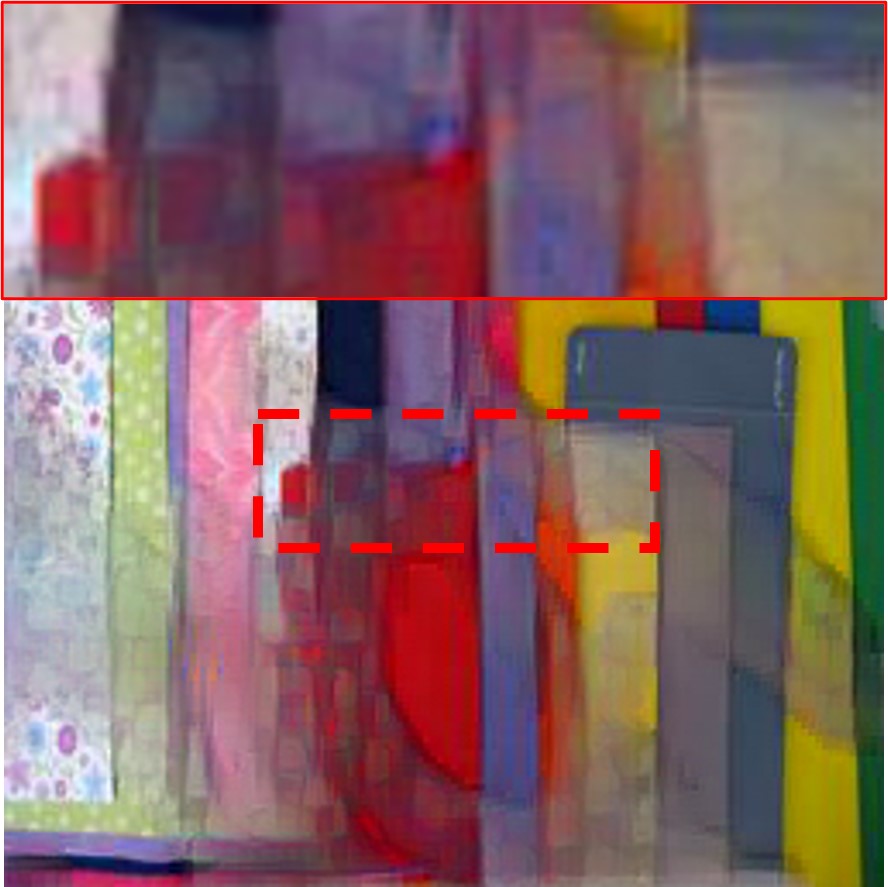}
    \includegraphics[width=0.1\textwidth]{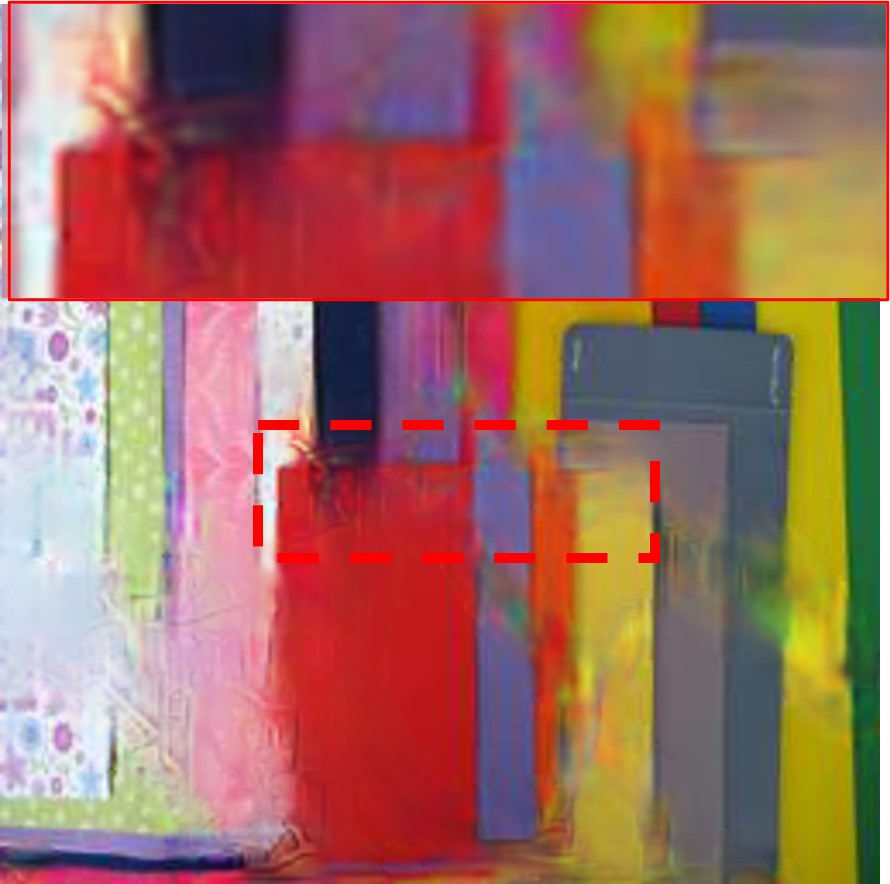}
    \includegraphics[width=0.1\textwidth]{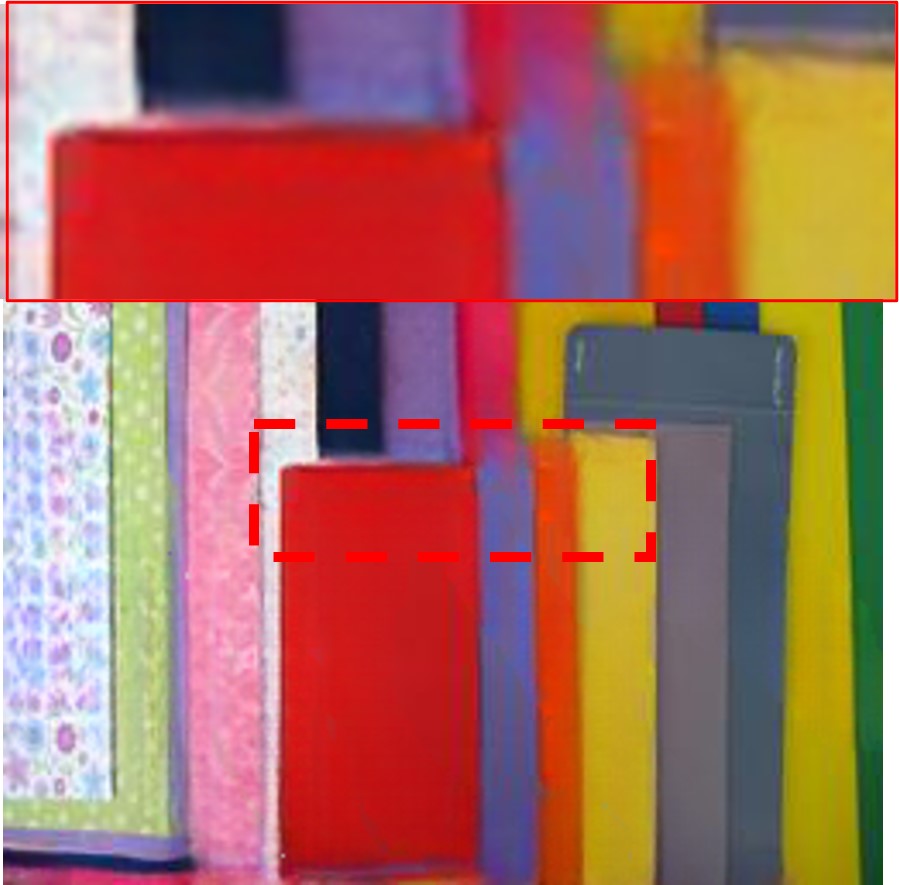}
    \includegraphics[width=0.1\textwidth]{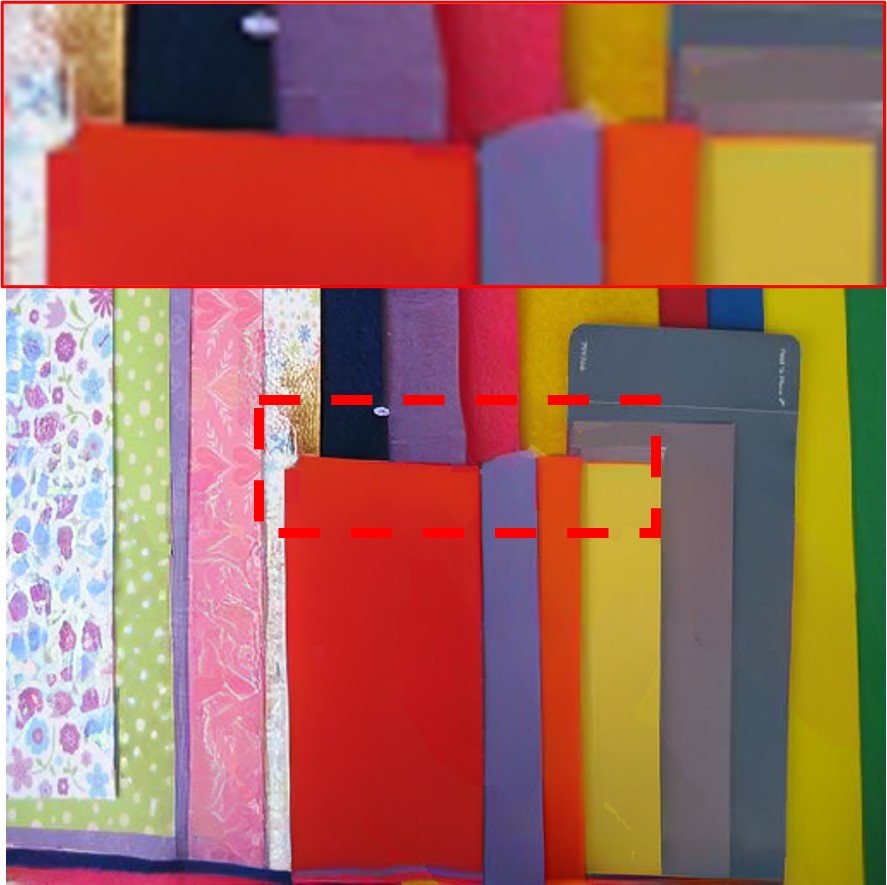}
    \includegraphics[width=0.1\textwidth]{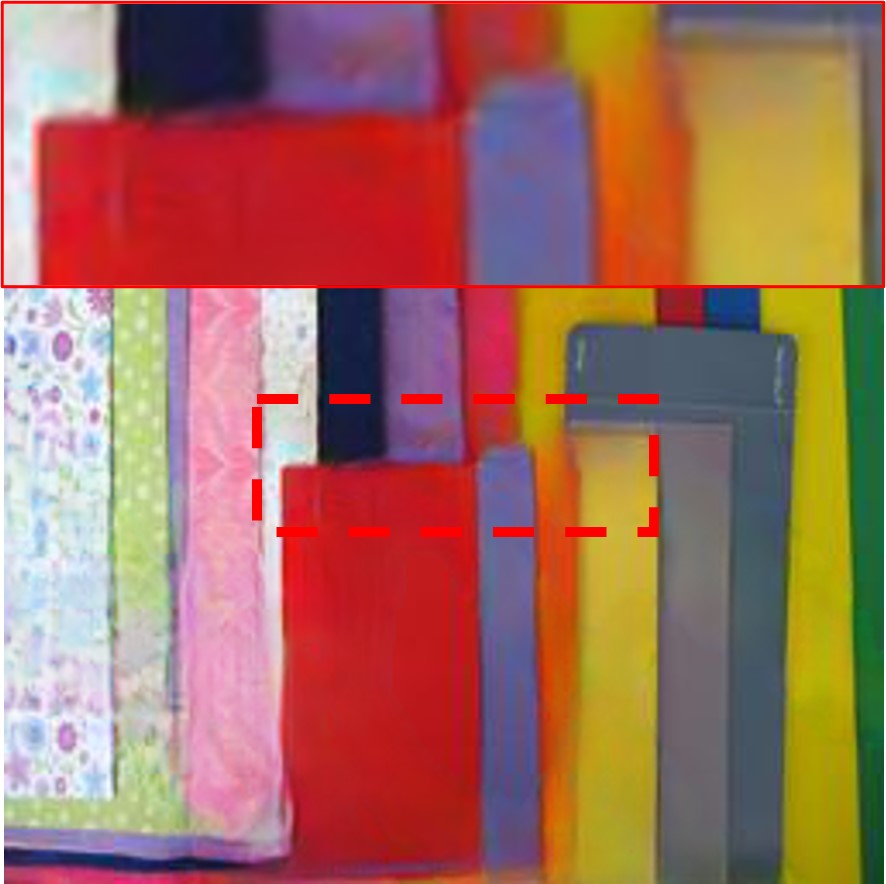}
    \includegraphics[width=0.1\textwidth]{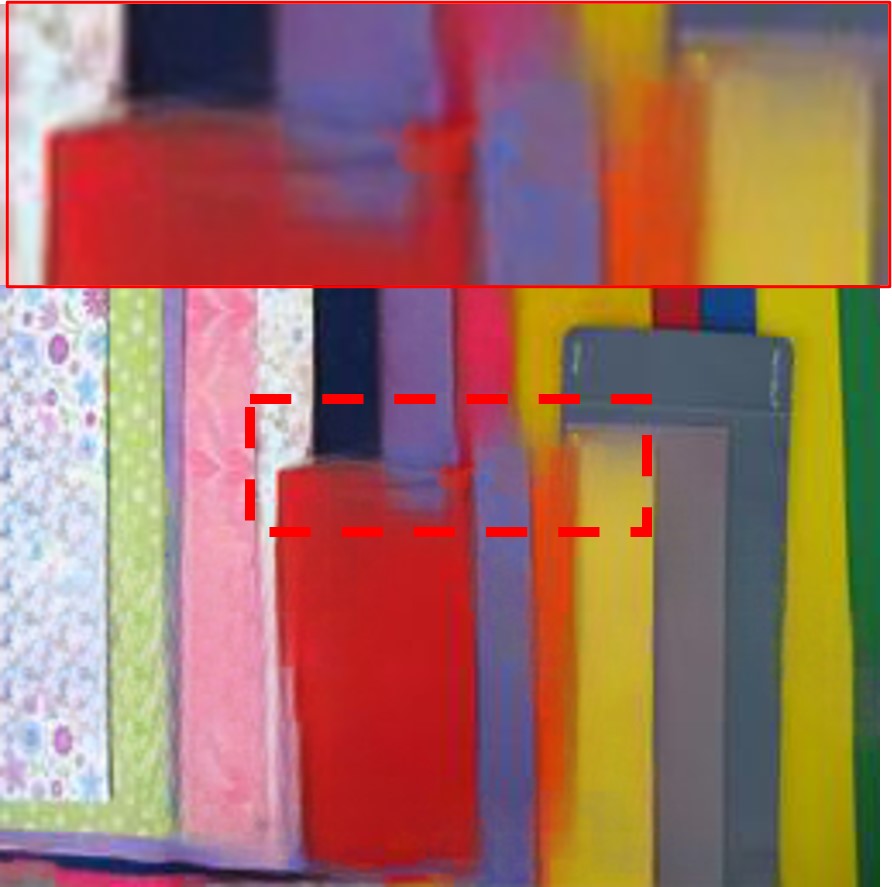}
    \includegraphics[width=0.1\textwidth]{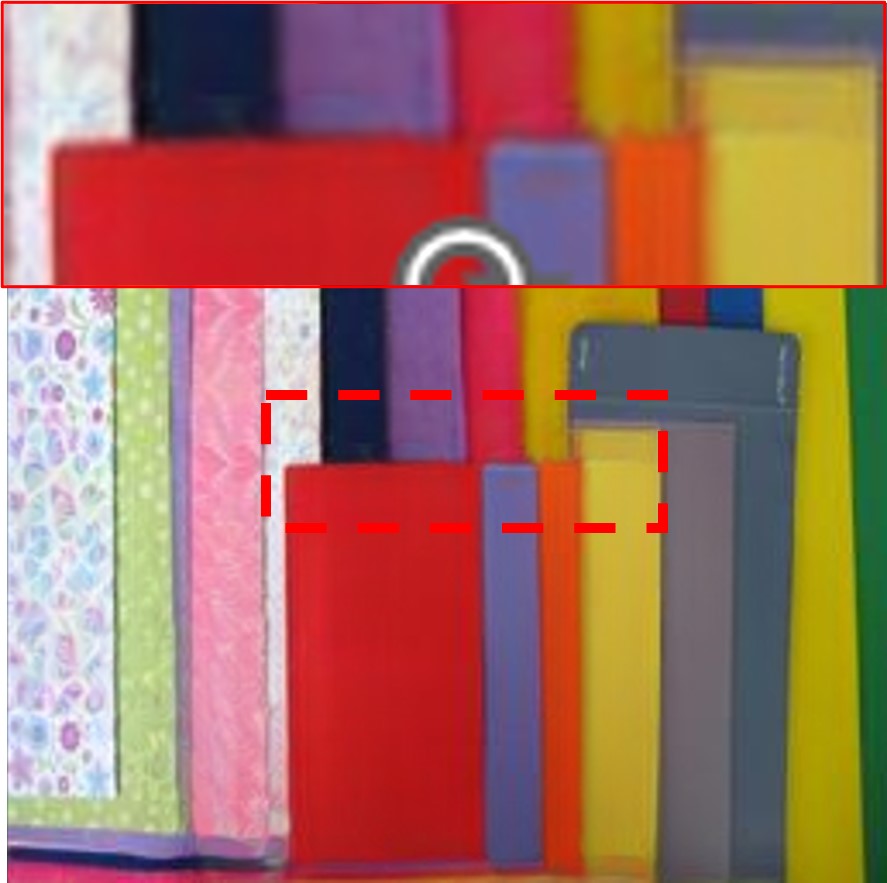}
    \includegraphics[width=0.1\textwidth]{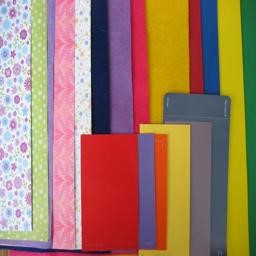}
    \\
    \includegraphics[width=0.1\textwidth]{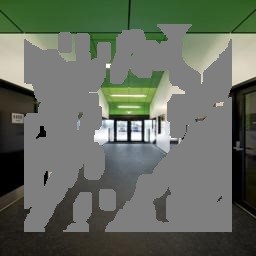}
    \includegraphics[width=0.1\textwidth]{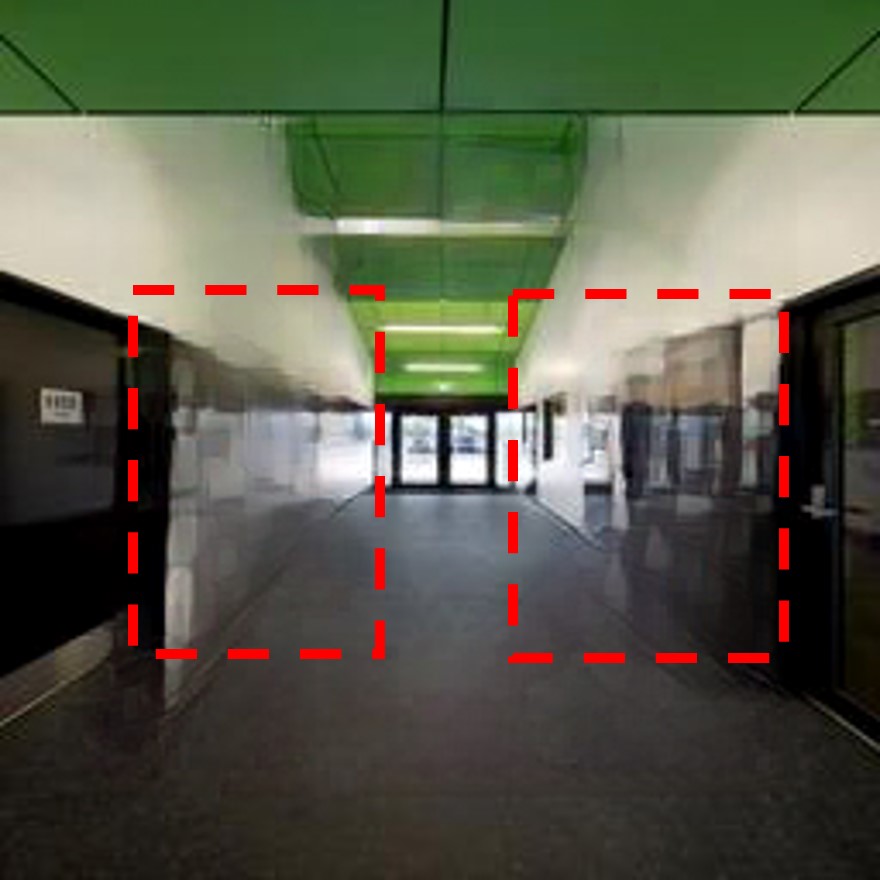}
    \includegraphics[width=0.1\textwidth]{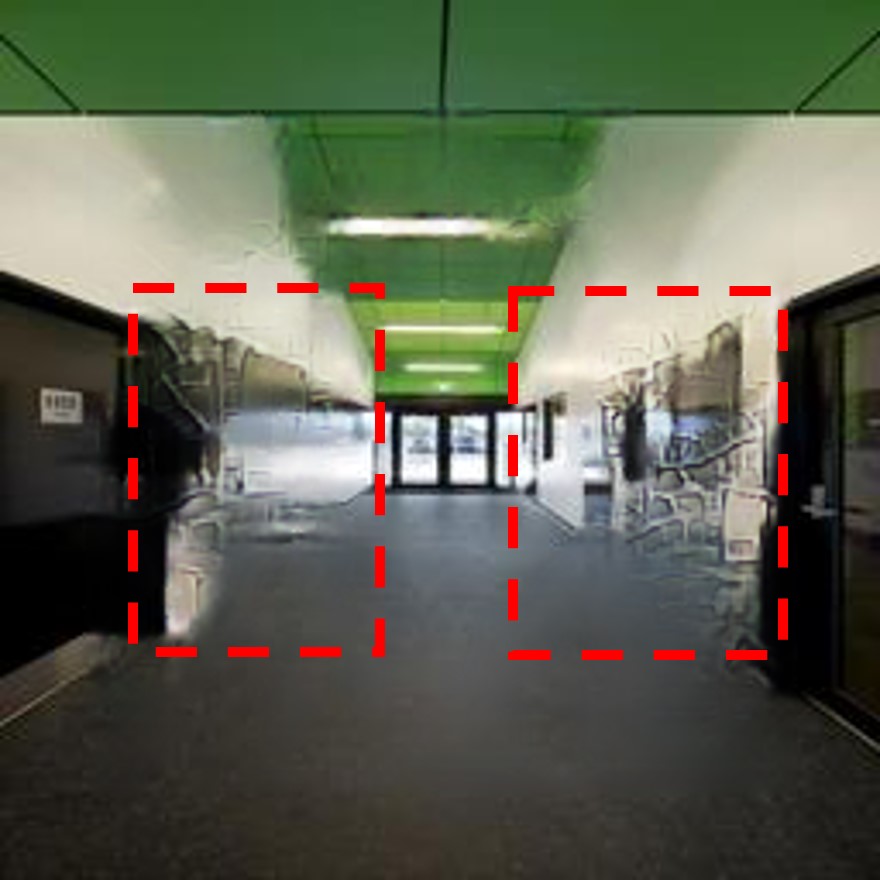}
    \includegraphics[width=0.1\textwidth]{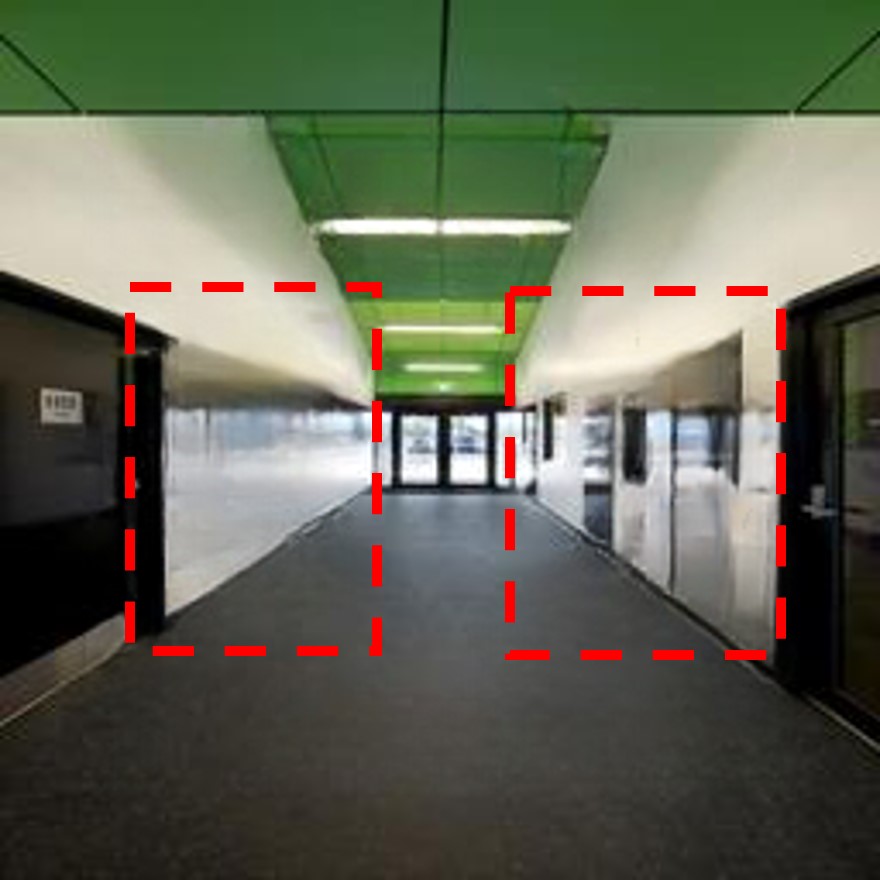}
    \includegraphics[width=0.1\textwidth]{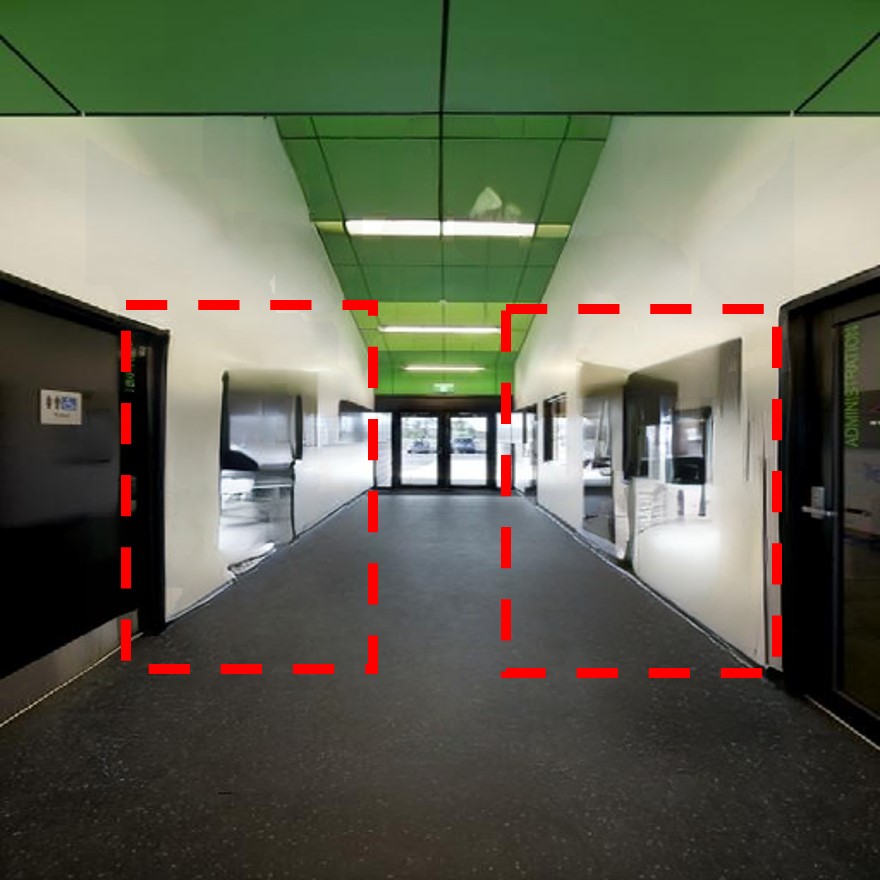}
    \includegraphics[width=0.1\textwidth]{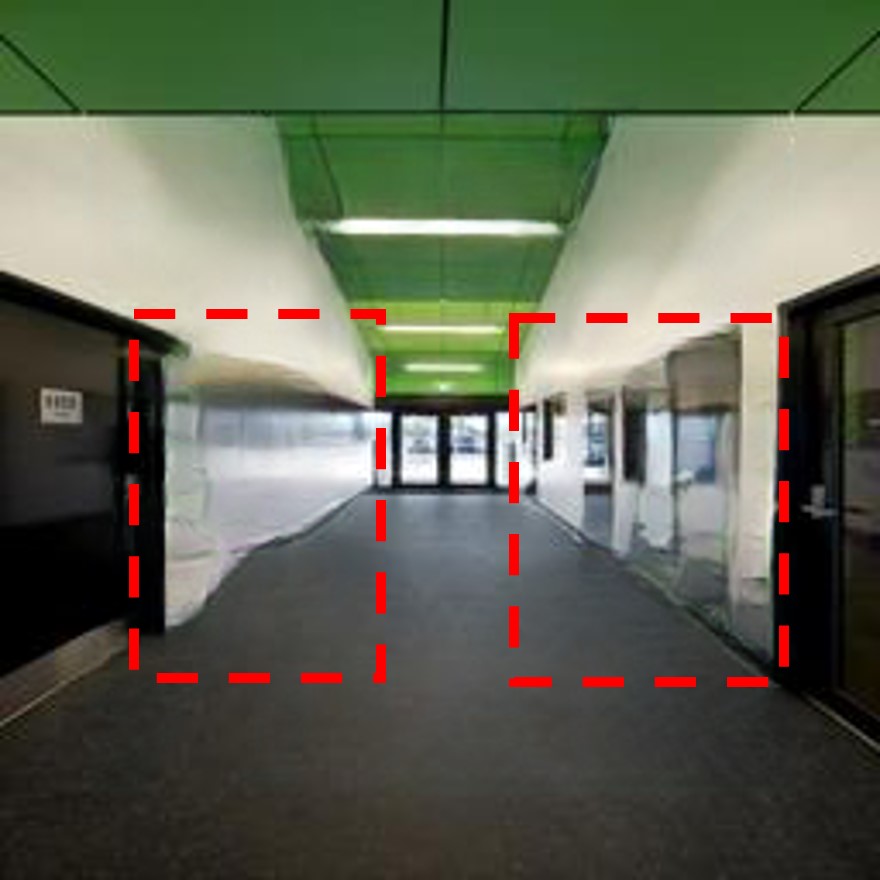}
    \includegraphics[width=0.1\textwidth]{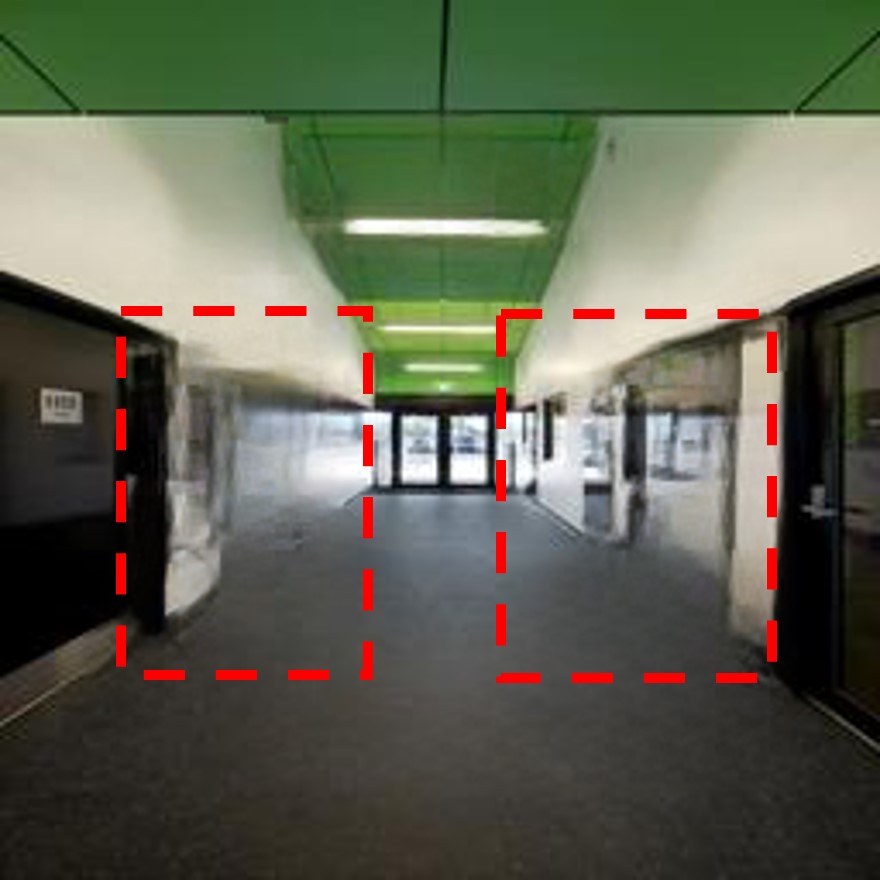}
    \includegraphics[width=0.1\textwidth]{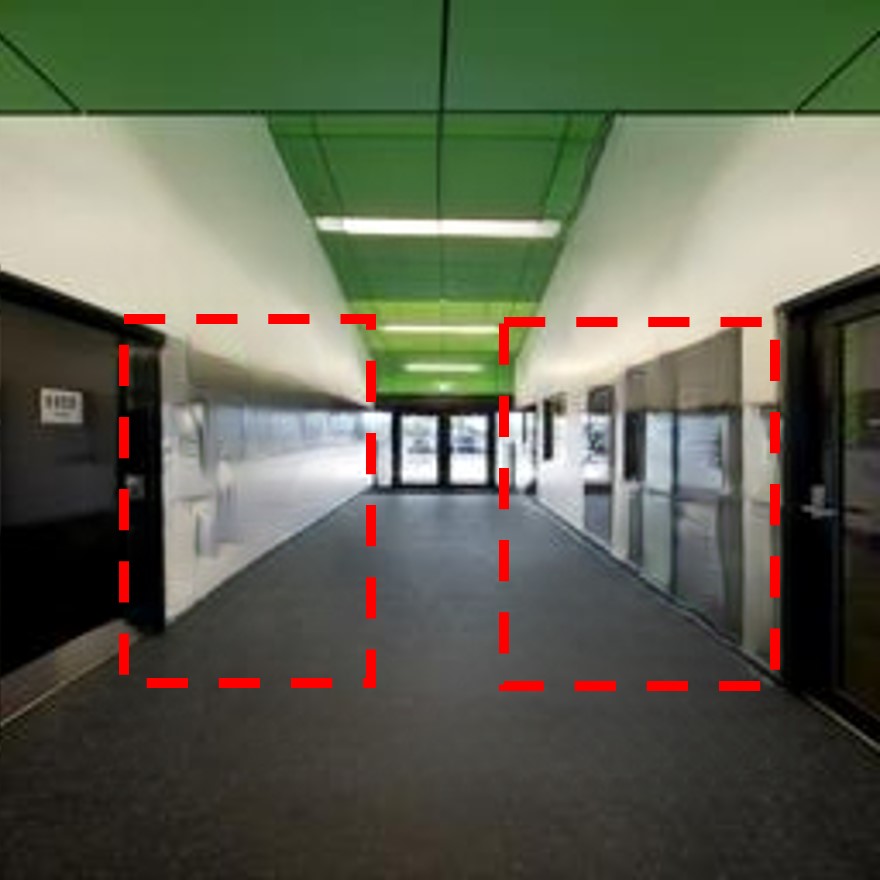}
    \includegraphics[width=0.1\textwidth]{img/places2_ano/gt2.jpg}
    \\
    \includegraphics[width=0.1\textwidth]{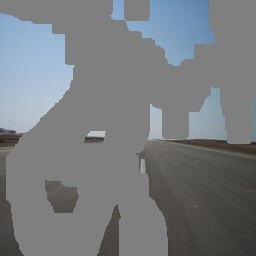}
    \includegraphics[width=0.1\textwidth]{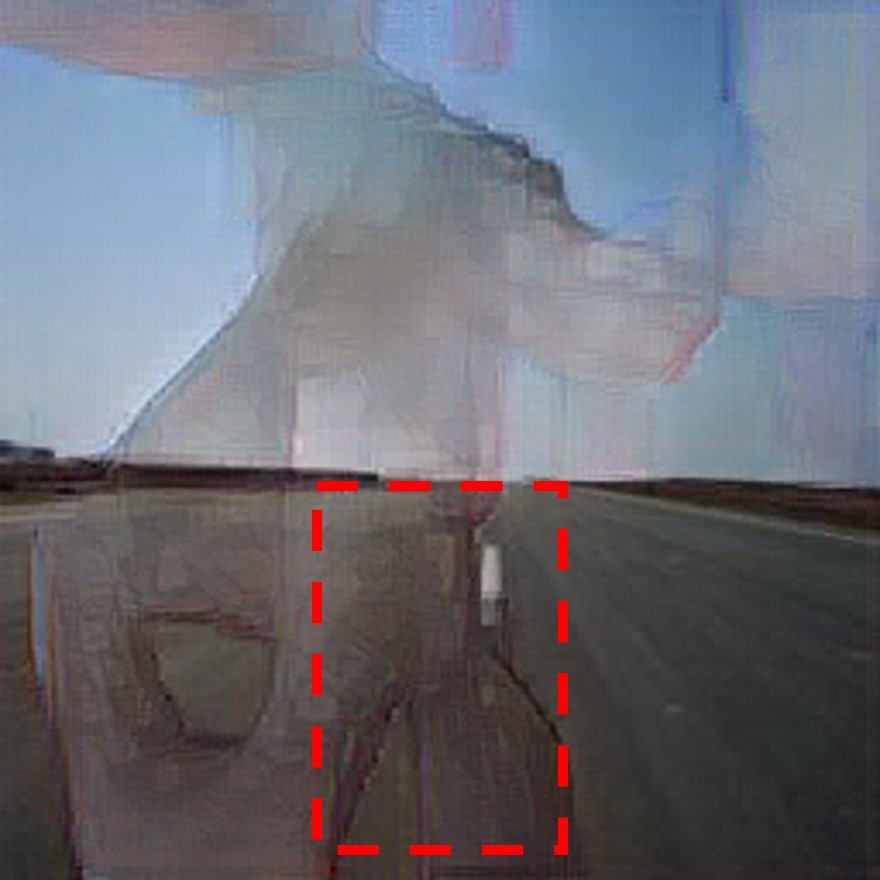}
    \includegraphics[width=0.1\textwidth]{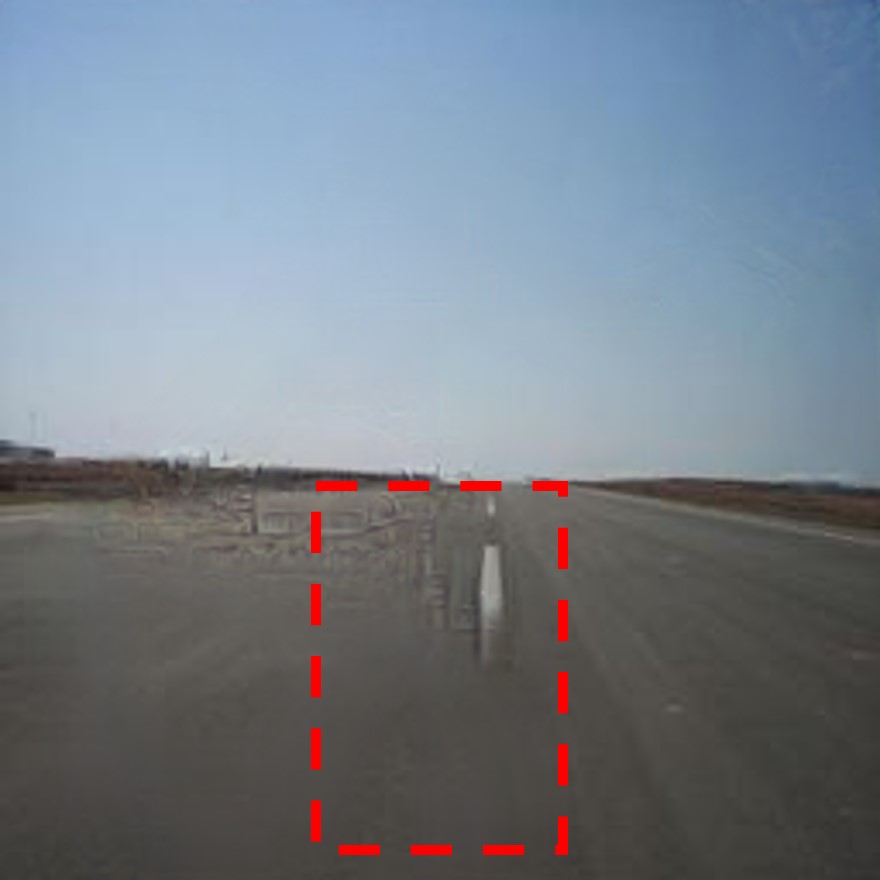}
    \includegraphics[width=0.1\textwidth]{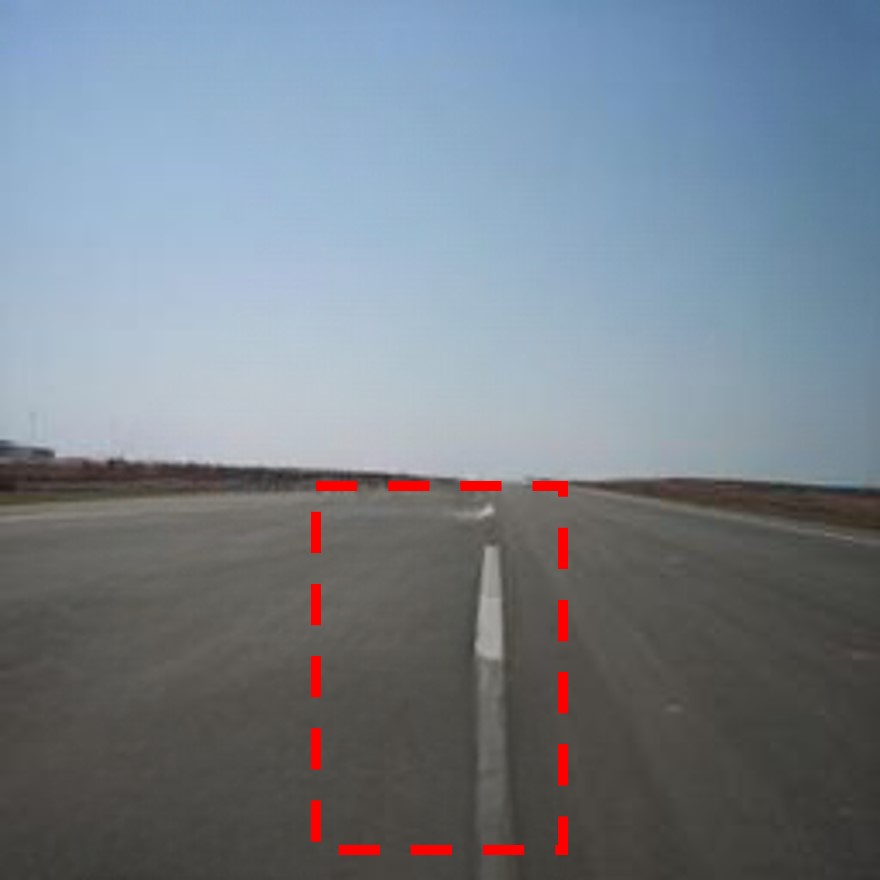}
    \includegraphics[width=0.1\textwidth]{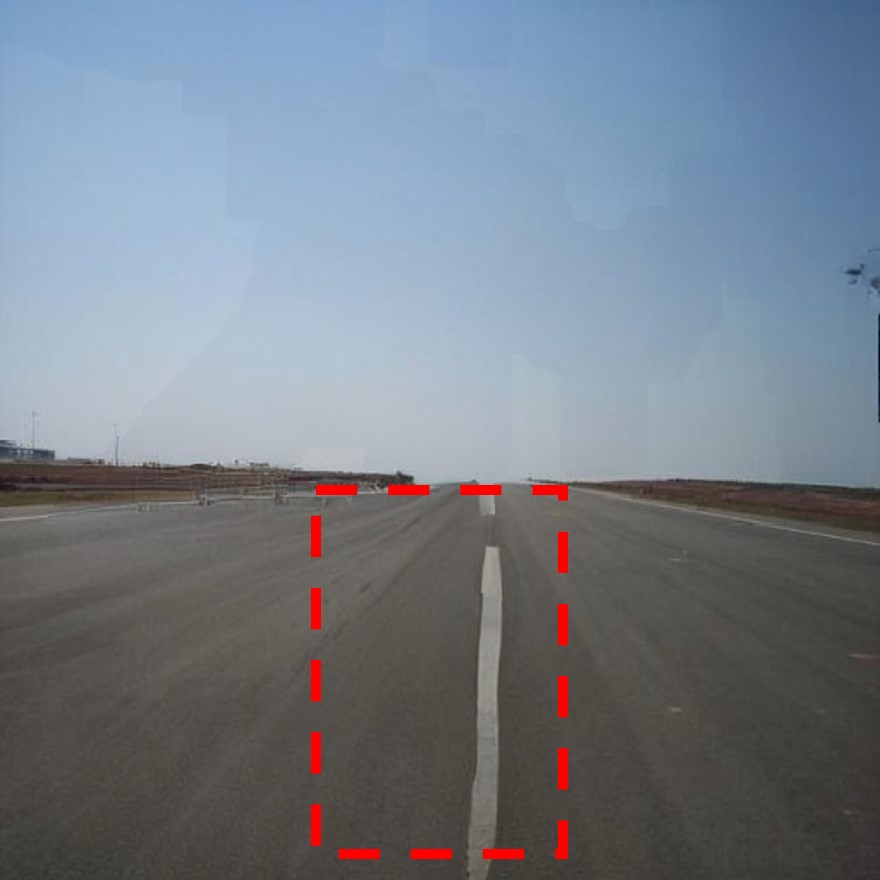}
    \includegraphics[width=0.1\textwidth]{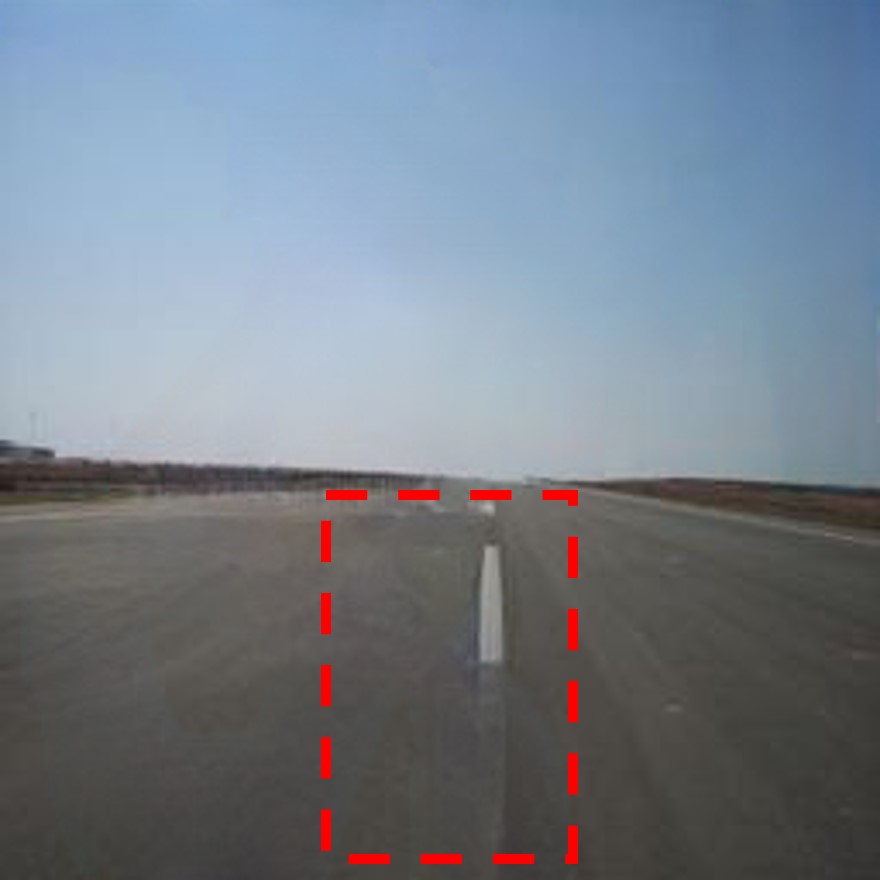}
    \includegraphics[width=0.1\textwidth]{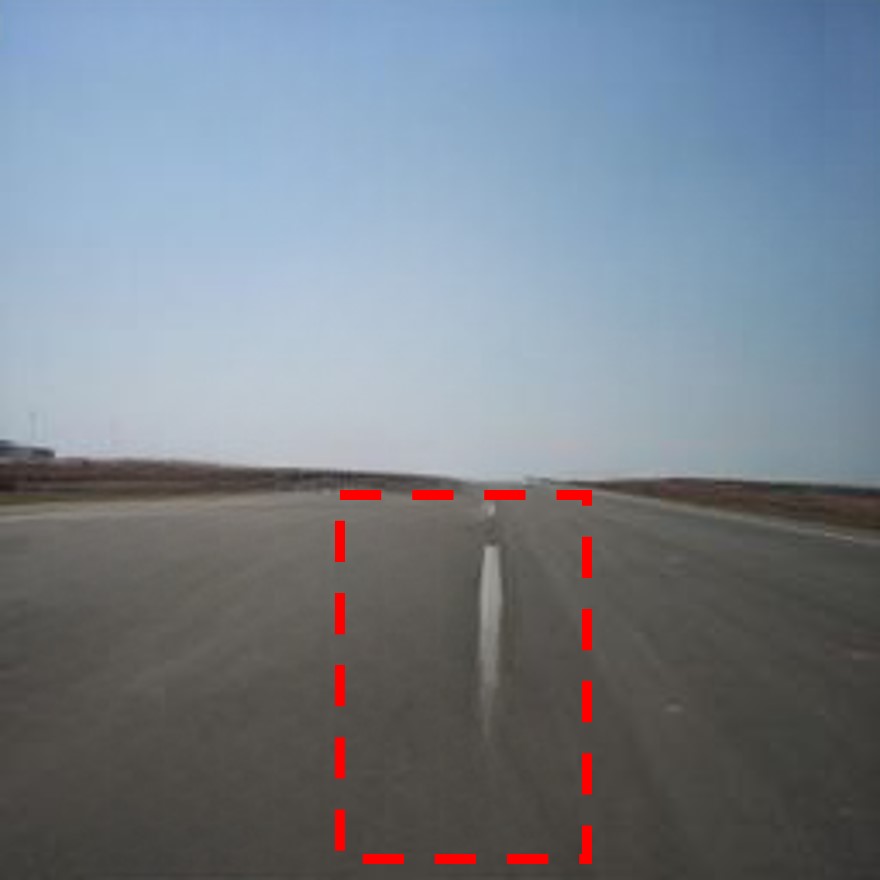}
    \includegraphics[width=0.1\textwidth]{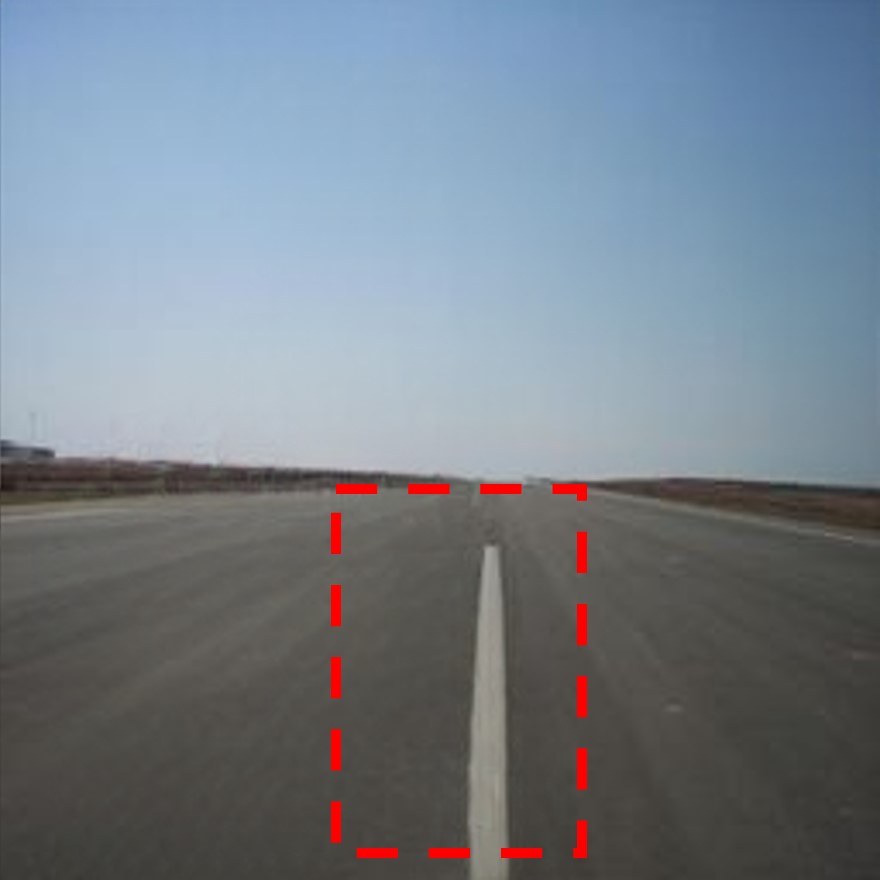}
    \includegraphics[width=0.1\textwidth]{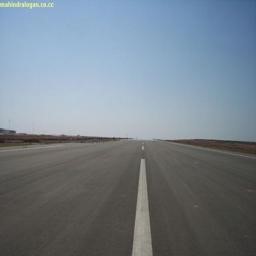}
    \\
    \makebox[0.1\textwidth]{\tiny Input}
    \makebox[0.1\textwidth]{\tiny WaveFill~\cite{yu2021wavefill}}
    \makebox[0.1\textwidth]{\tiny CTSDG~\cite{guo2021image}}
    \makebox[0.1\textwidth]{\tiny LAMA~\cite{wan2021high}}
    \makebox[0.1\textwidth]{\tiny LDM~\cite{Rombach_2022_CVPR}}
    \makebox[0.1\textwidth]{\tiny MISF~\cite{li2022misf}}
    \makebox[0.1\textwidth]{\tiny CMT~\cite{ko2023continuously}}
    \makebox[0.1\textwidth]{\tiny Ours}
    \makebox[0.1\textwidth]{\tiny GT}
    \\
    \includegraphics[width=0.1\textwidth]{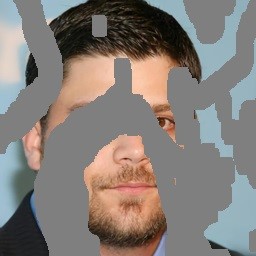}
    \includegraphics[width=0.1\textwidth]{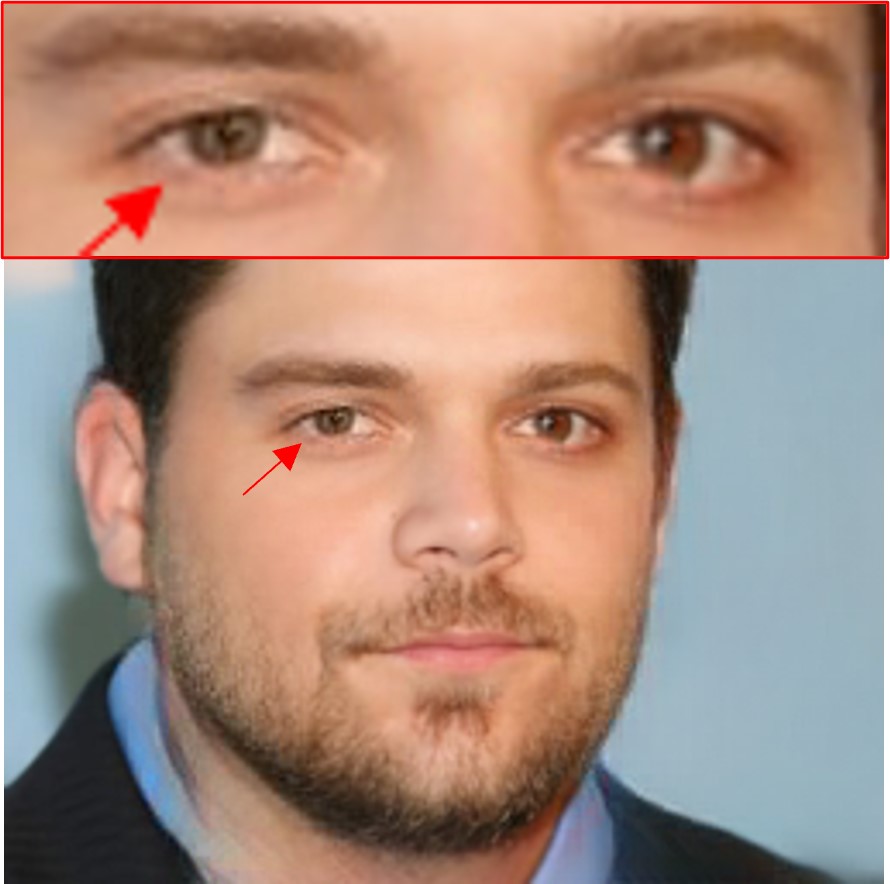}
    \includegraphics[width=0.1\textwidth]{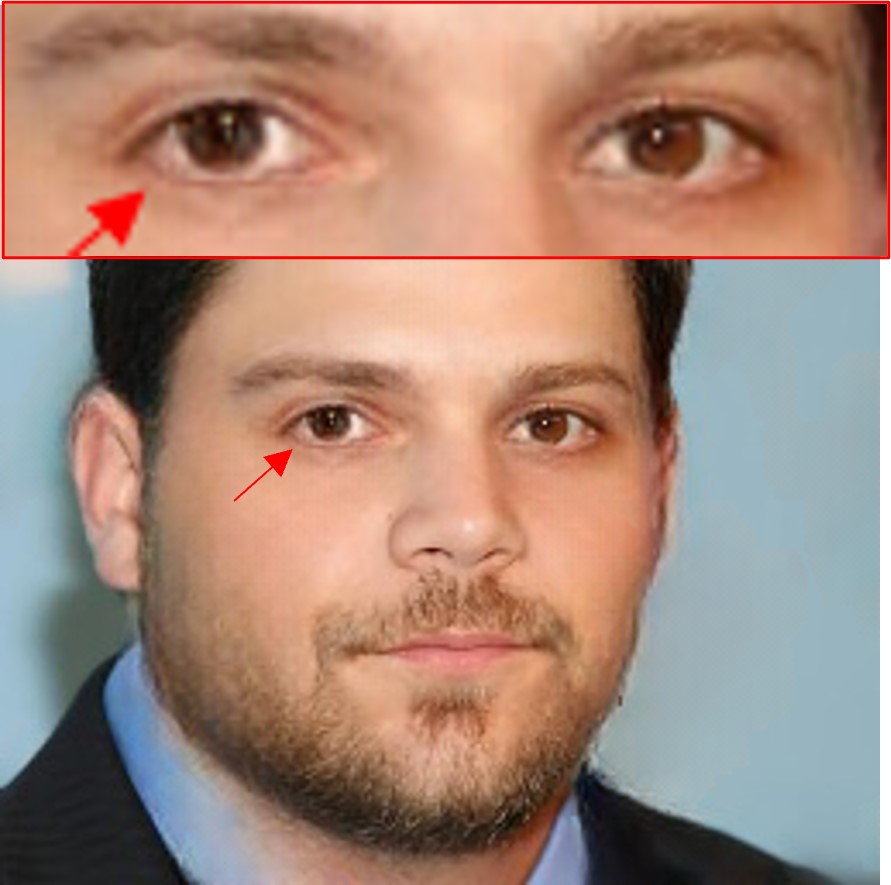}
    \includegraphics[width=0.1\textwidth]{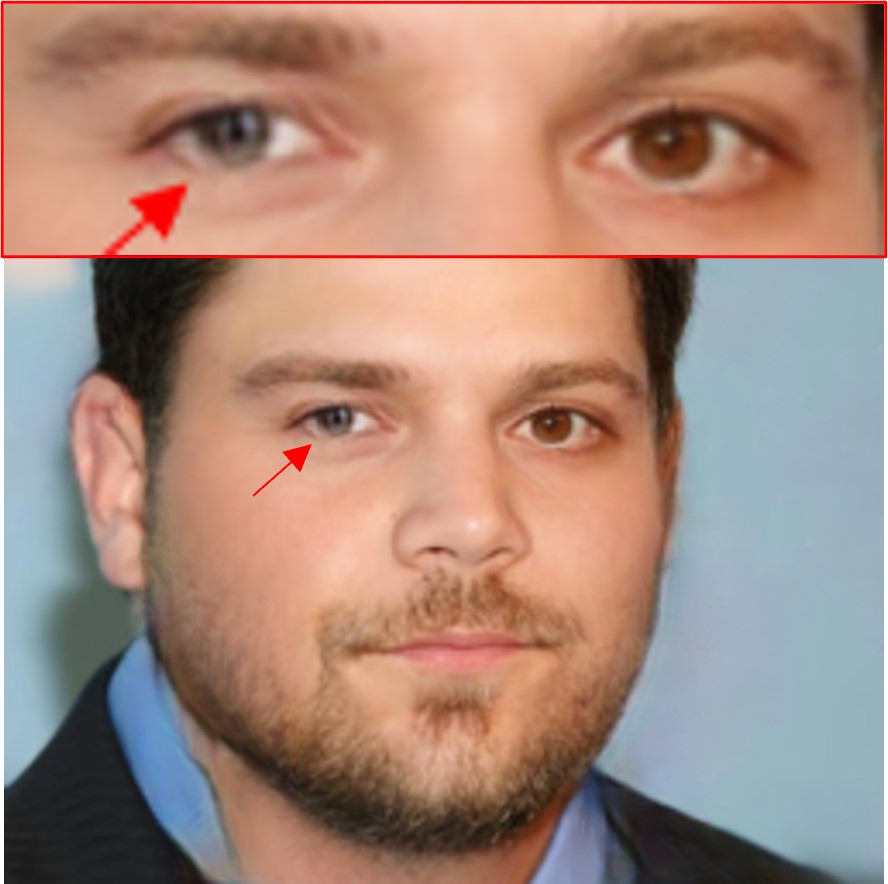}
    \includegraphics[width=0.1\textwidth]{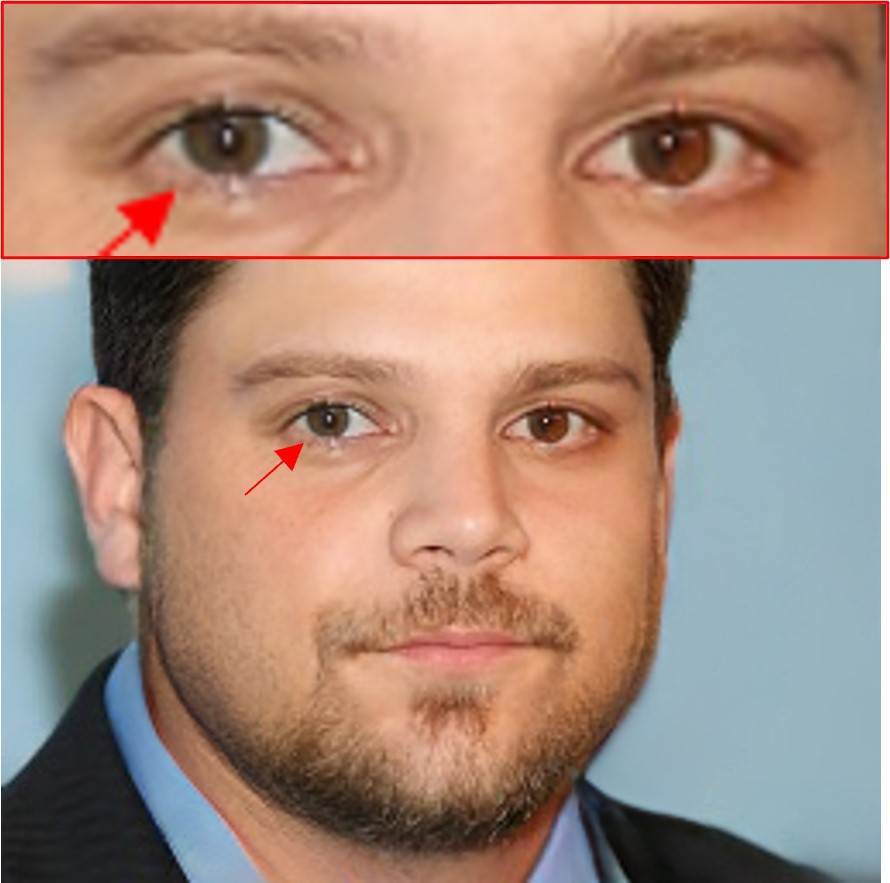}
    \includegraphics[width=0.1\textwidth]{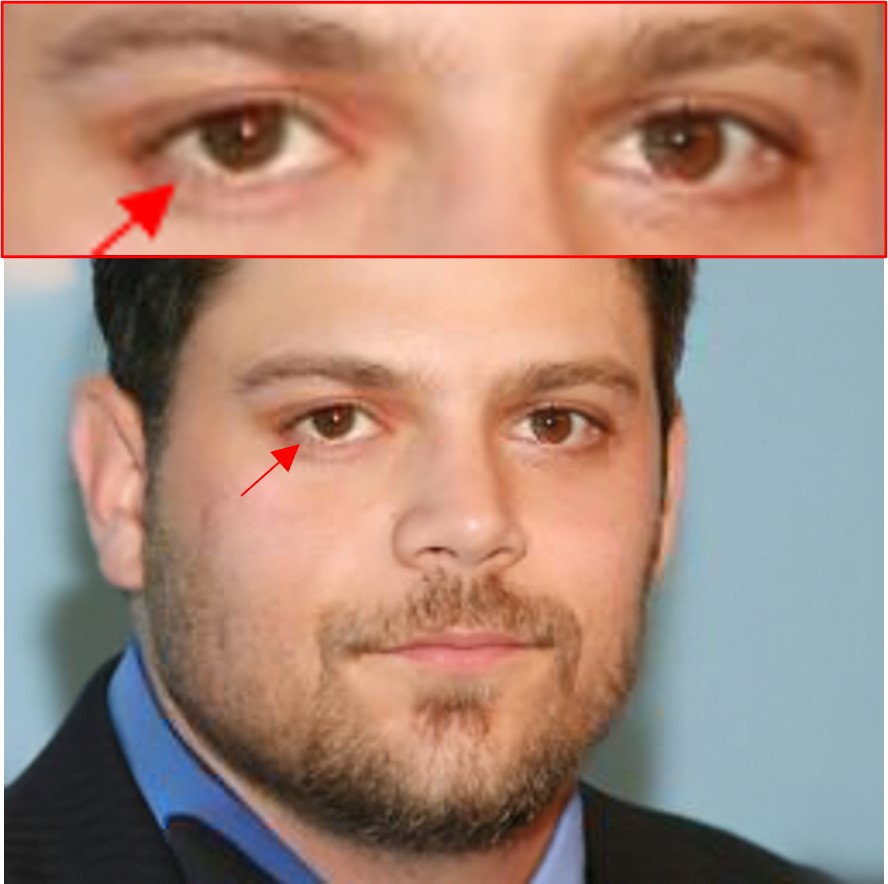}
    \includegraphics[width=0.1\textwidth]{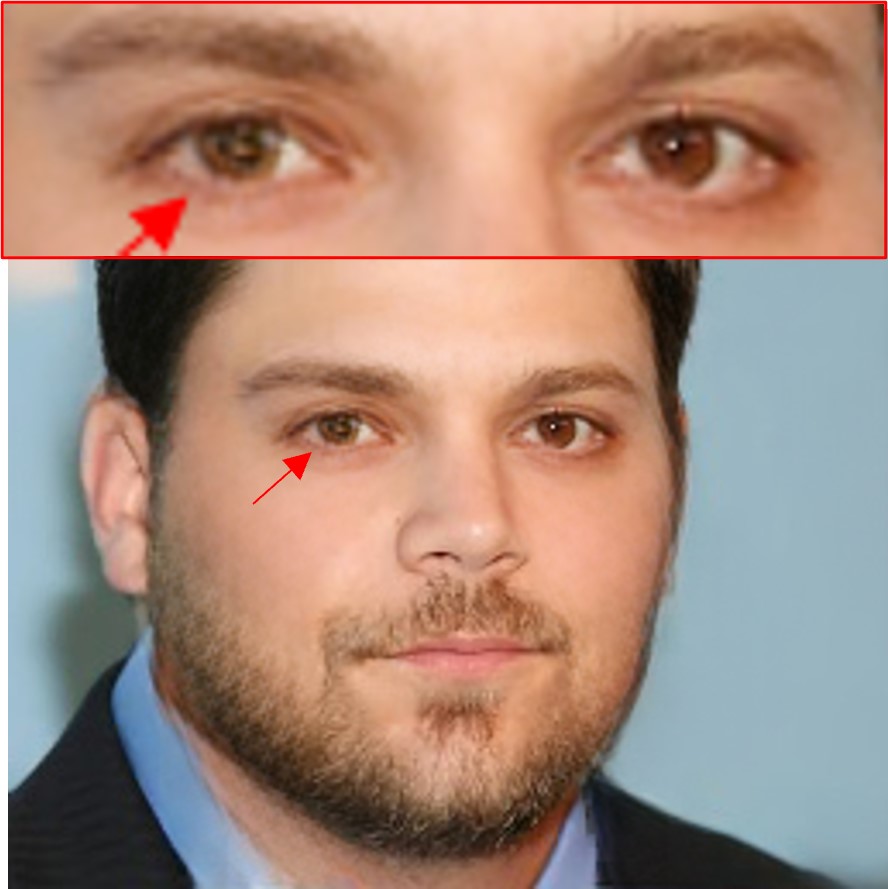}
    \includegraphics[width=0.1\textwidth]{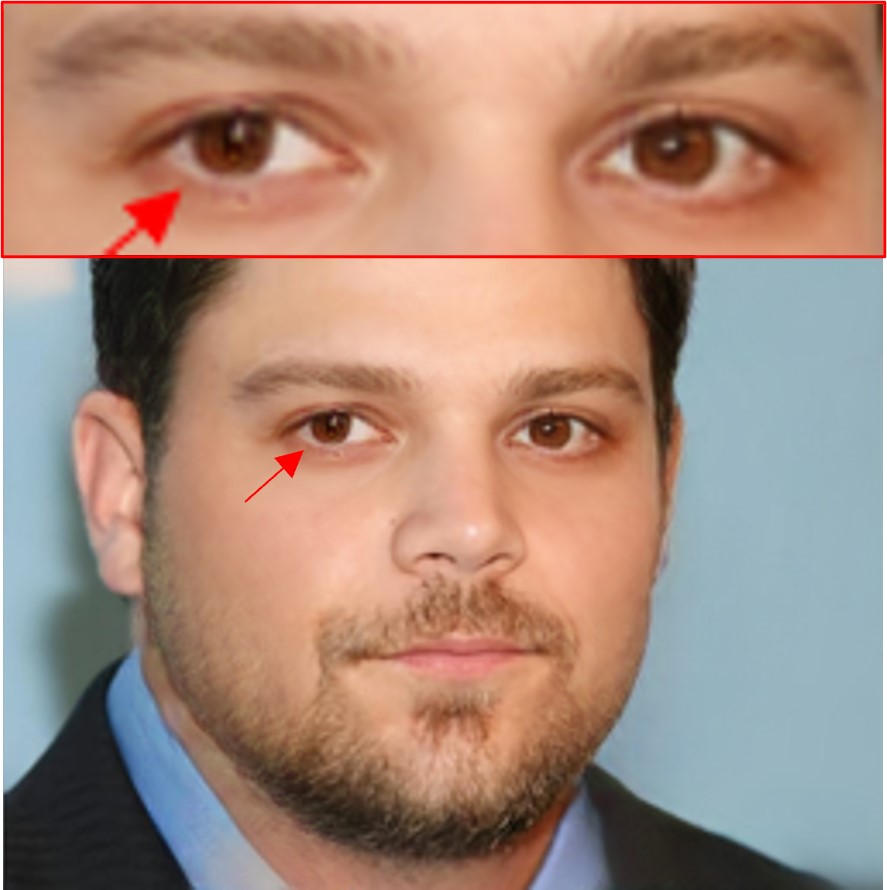}
    \includegraphics[width=0.1\textwidth]{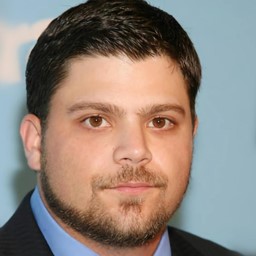}
    \\
    \includegraphics[width=0.1\textwidth]{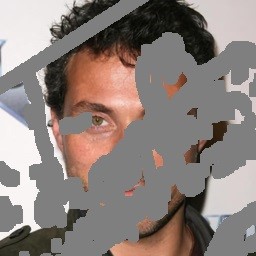}
    \includegraphics[width=0.1\textwidth]{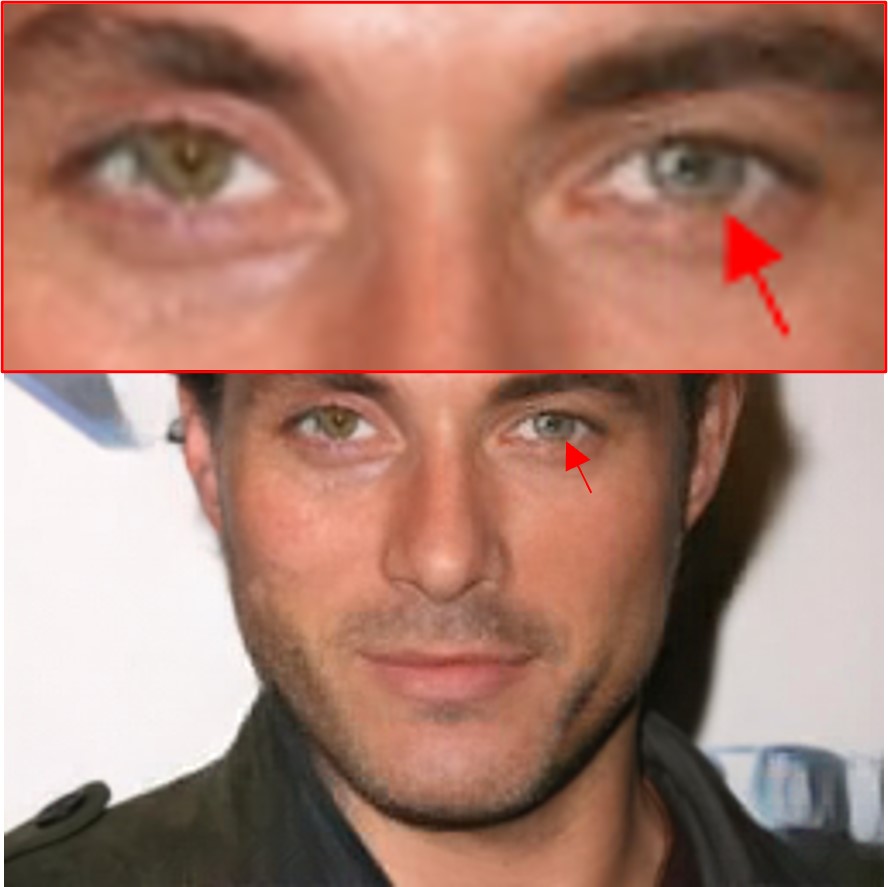}
    \includegraphics[width=0.1\textwidth]{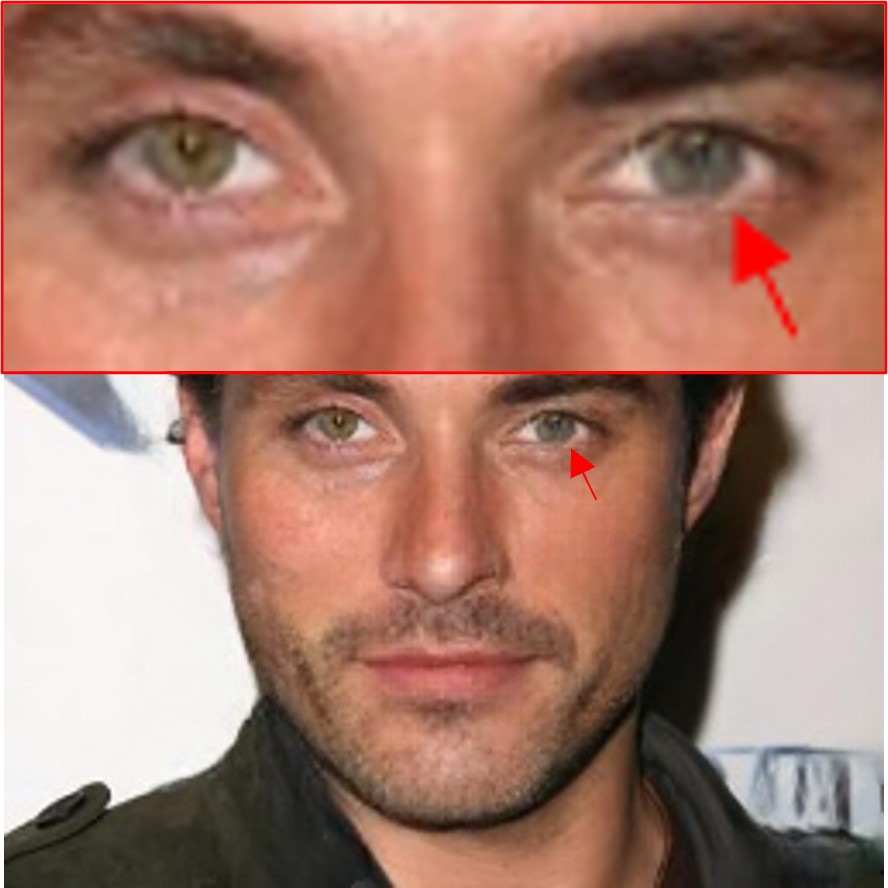}
    \includegraphics[width=0.1\textwidth]{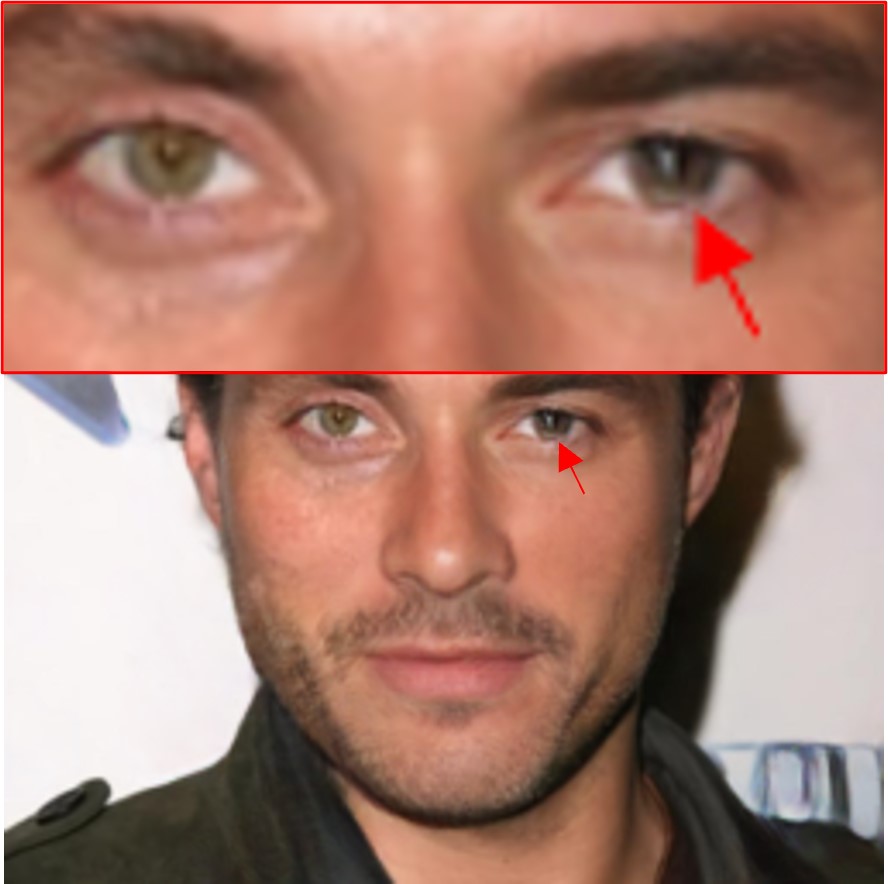}
    \includegraphics[width=0.1\textwidth]{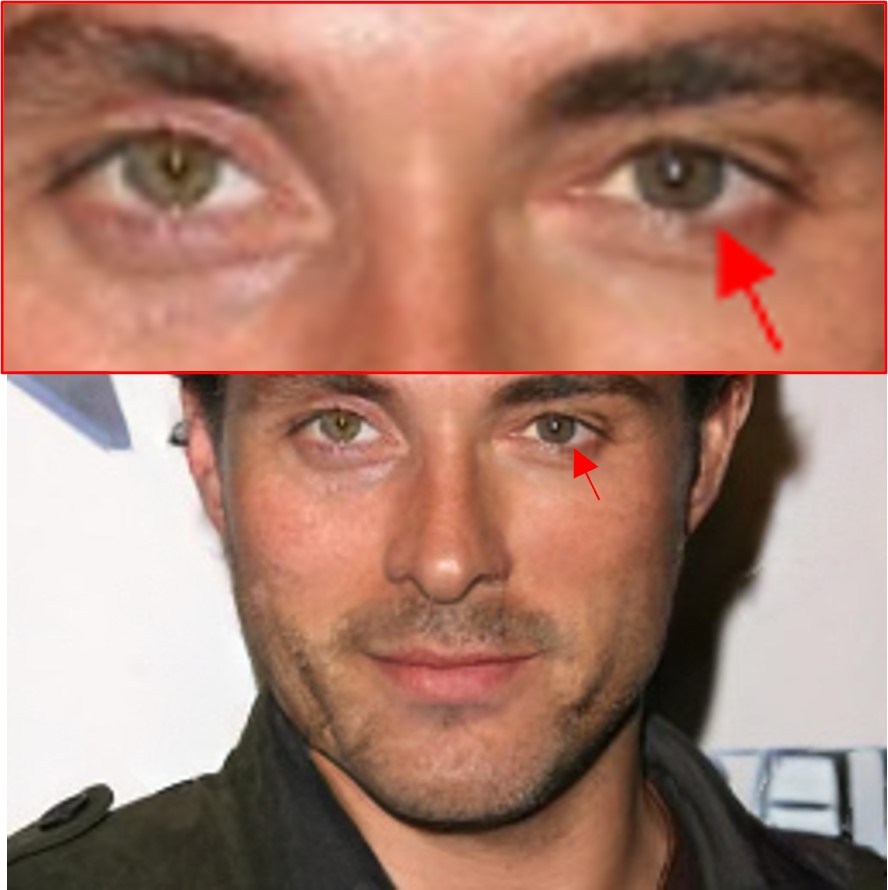}
    \includegraphics[width=0.1\textwidth]{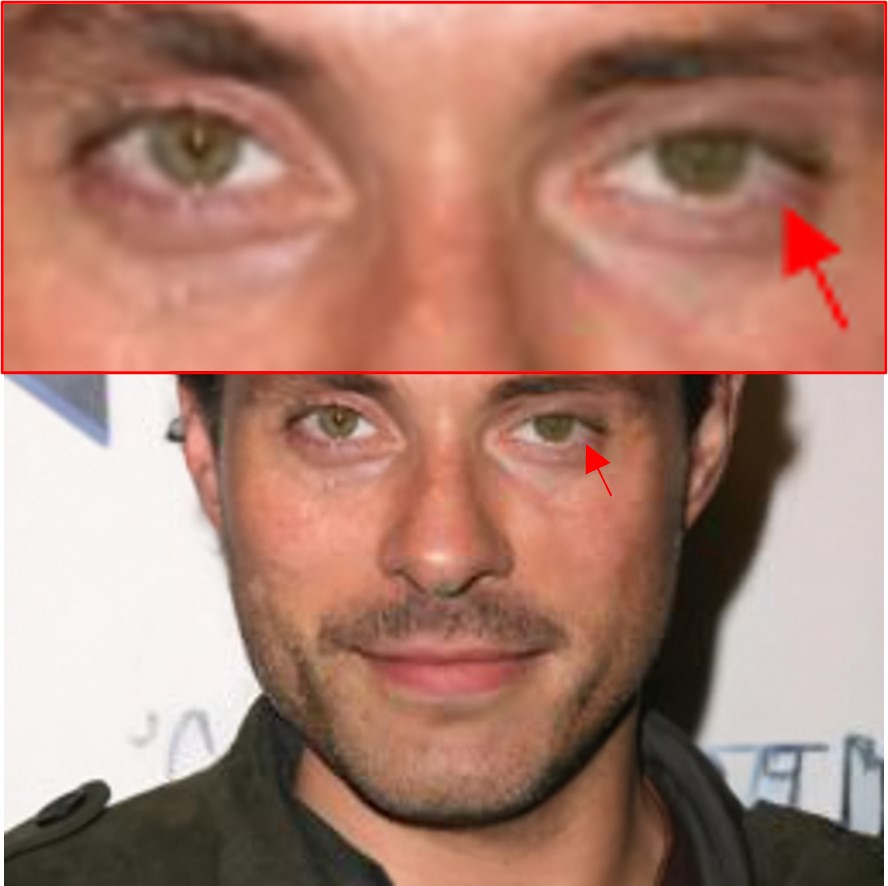}
    \includegraphics[width=0.1\textwidth]{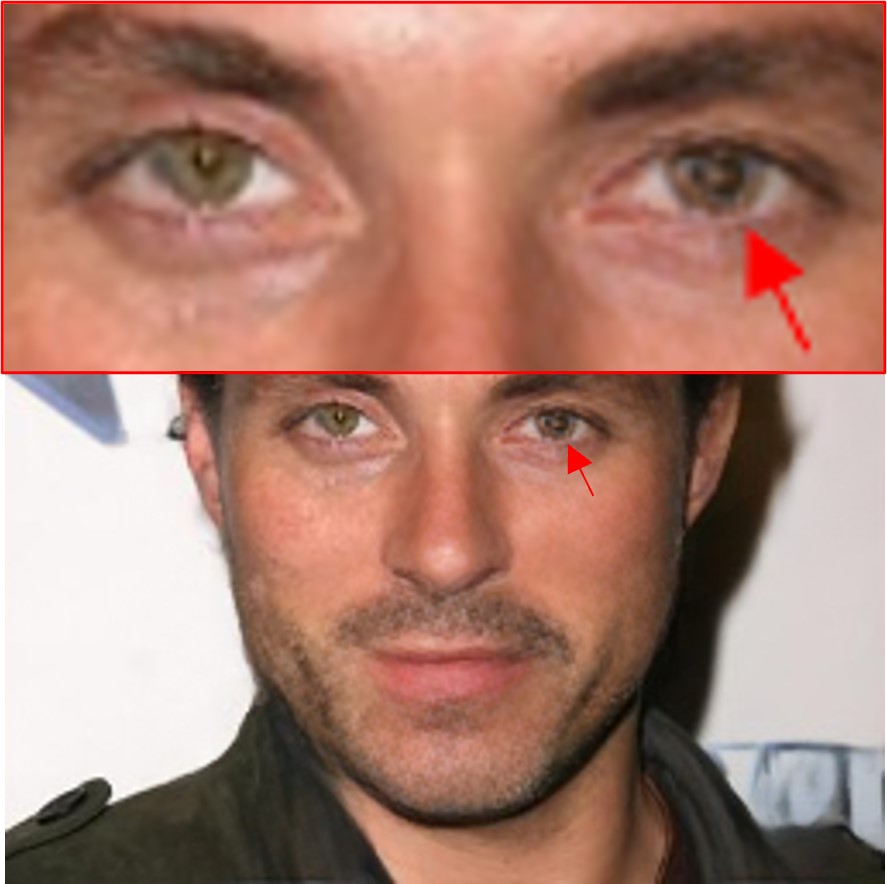}
    \includegraphics[width=0.1\textwidth]{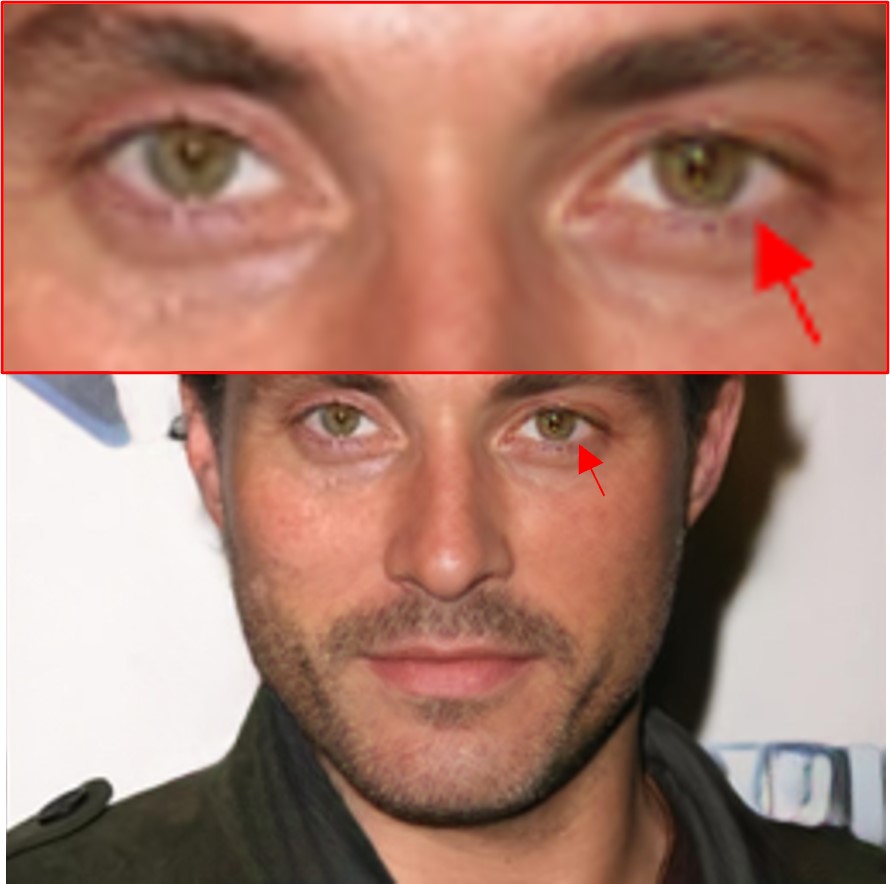}
    \includegraphics[width=0.1\textwidth]{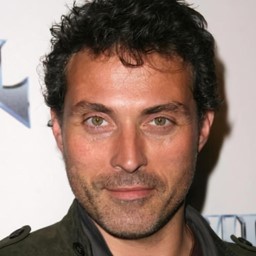}
    \\
    \includegraphics[width=0.1\textwidth]{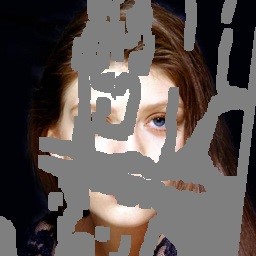}
    \includegraphics[width=0.1\textwidth]{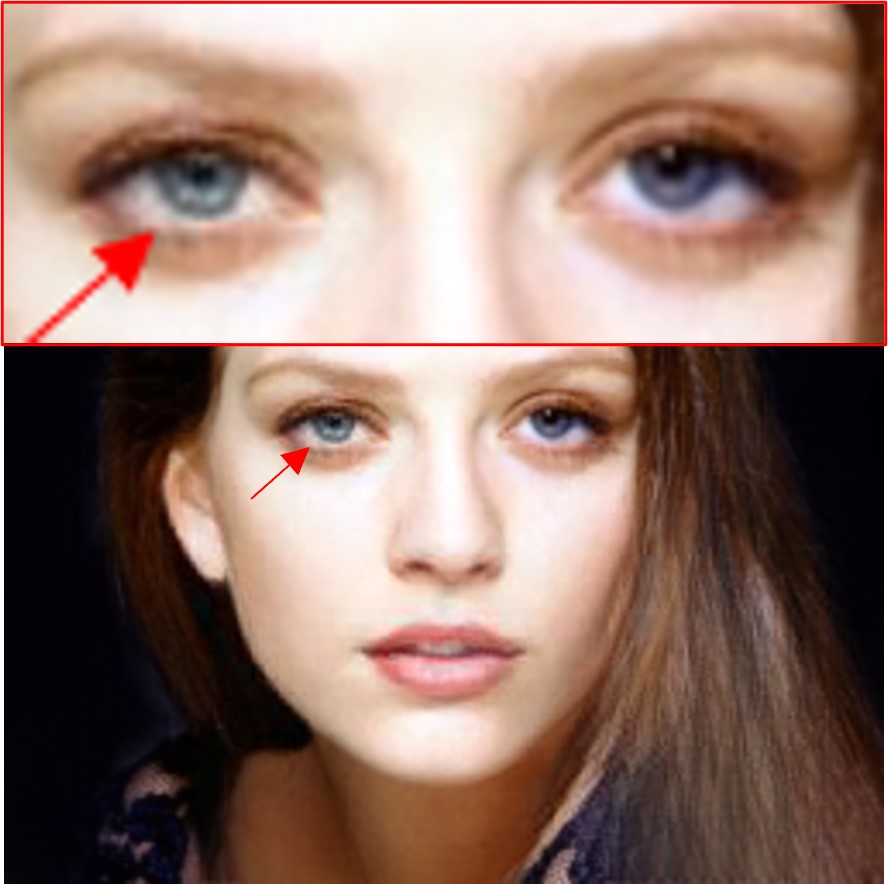}
    \includegraphics[width=0.1\textwidth]{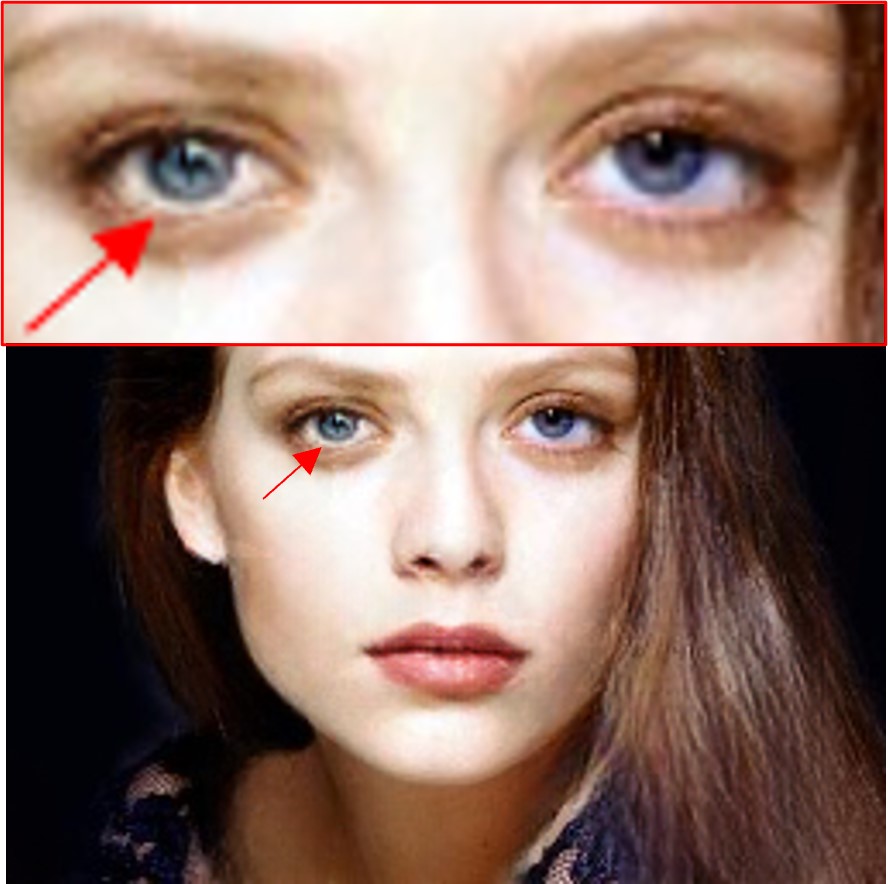}
    \includegraphics[width=0.1\textwidth]{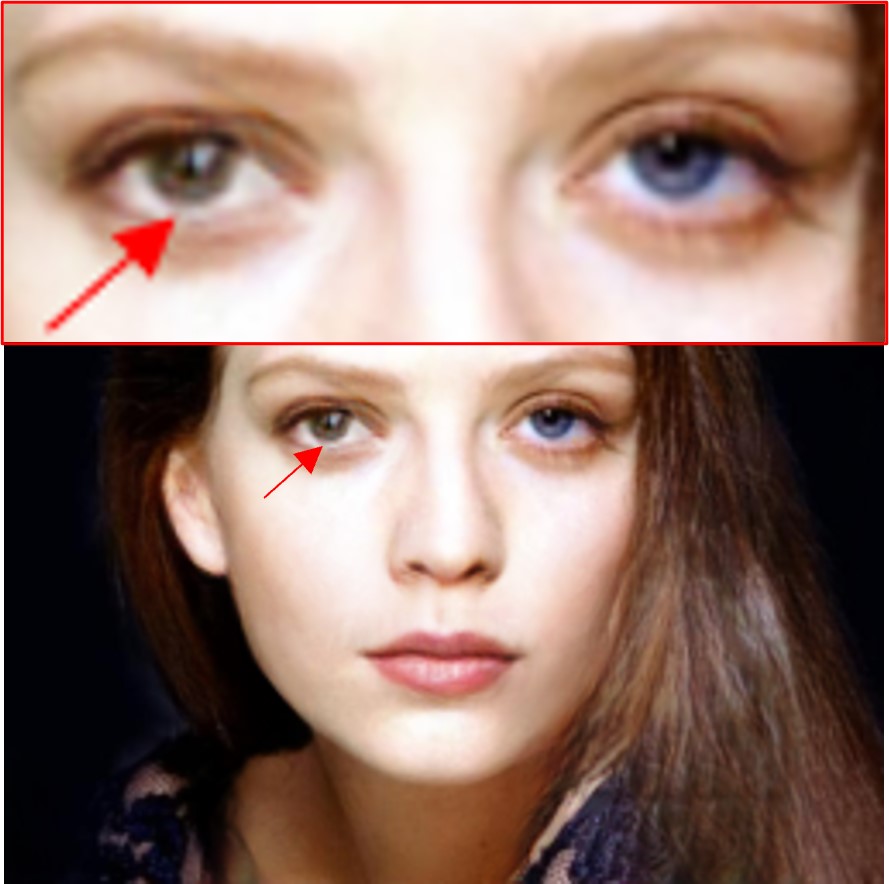}
    \includegraphics[width=0.1\textwidth]{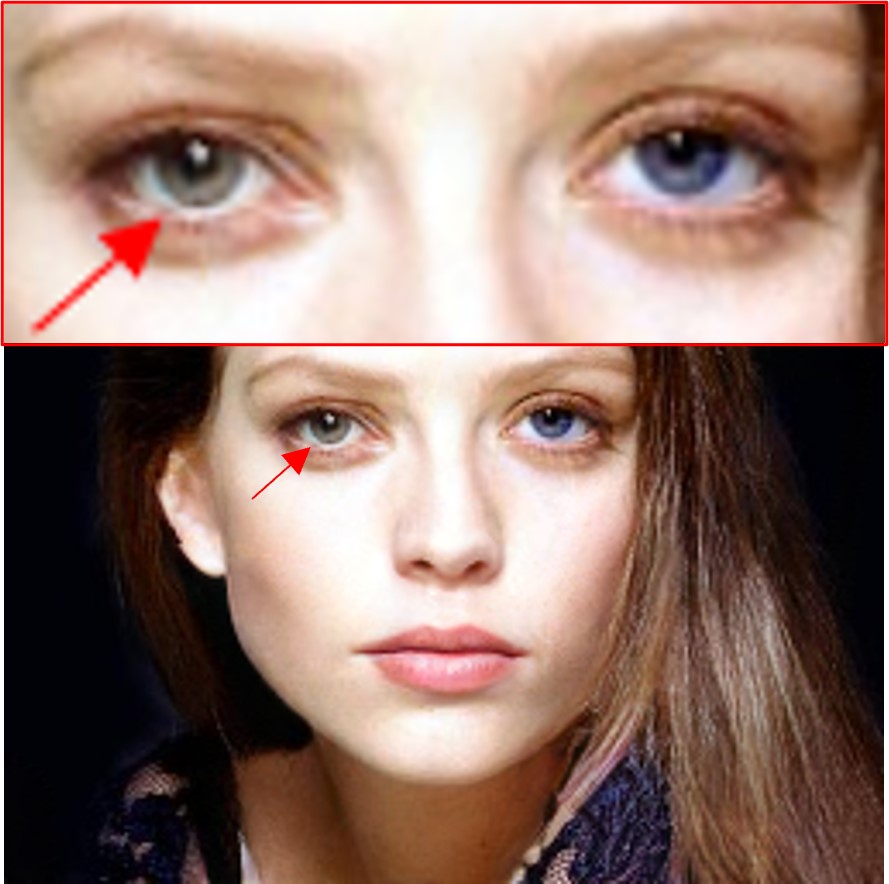}
    \includegraphics[width=0.1\textwidth]{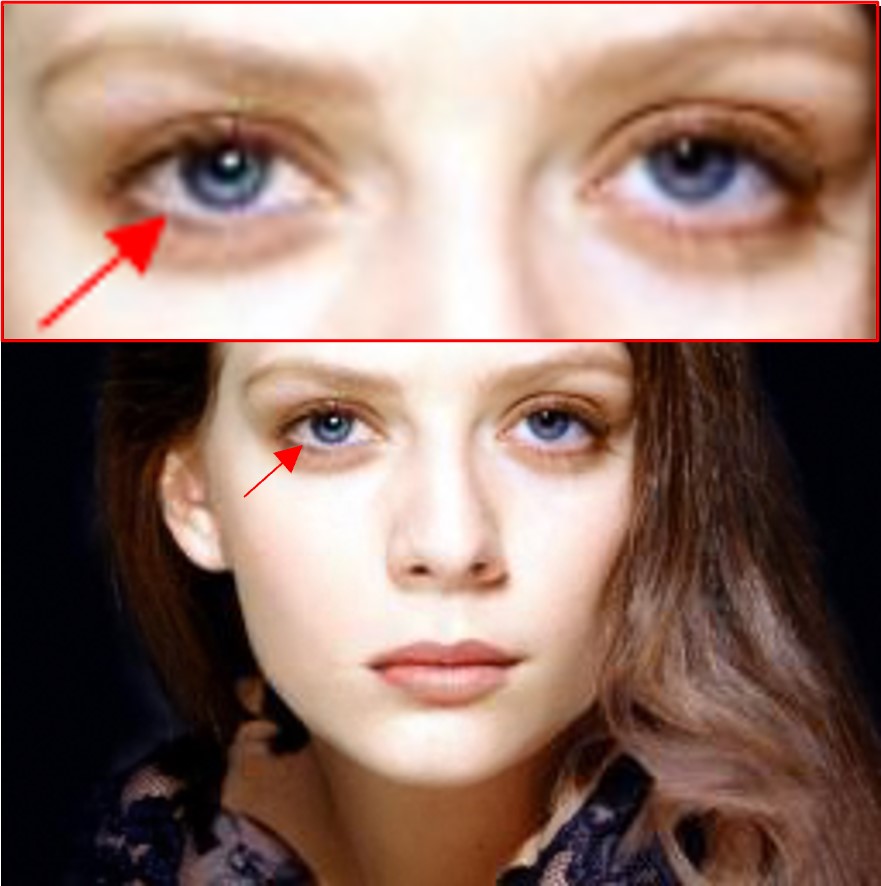}
    \includegraphics[width=0.1\textwidth]{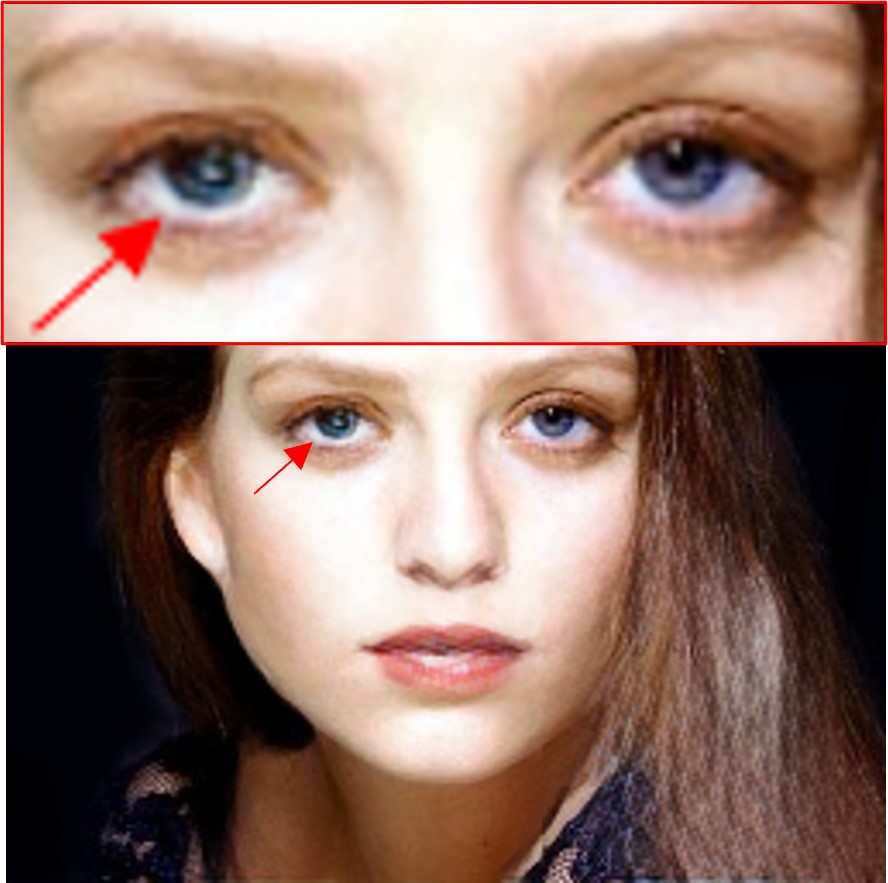}
    \includegraphics[width=0.1\textwidth]{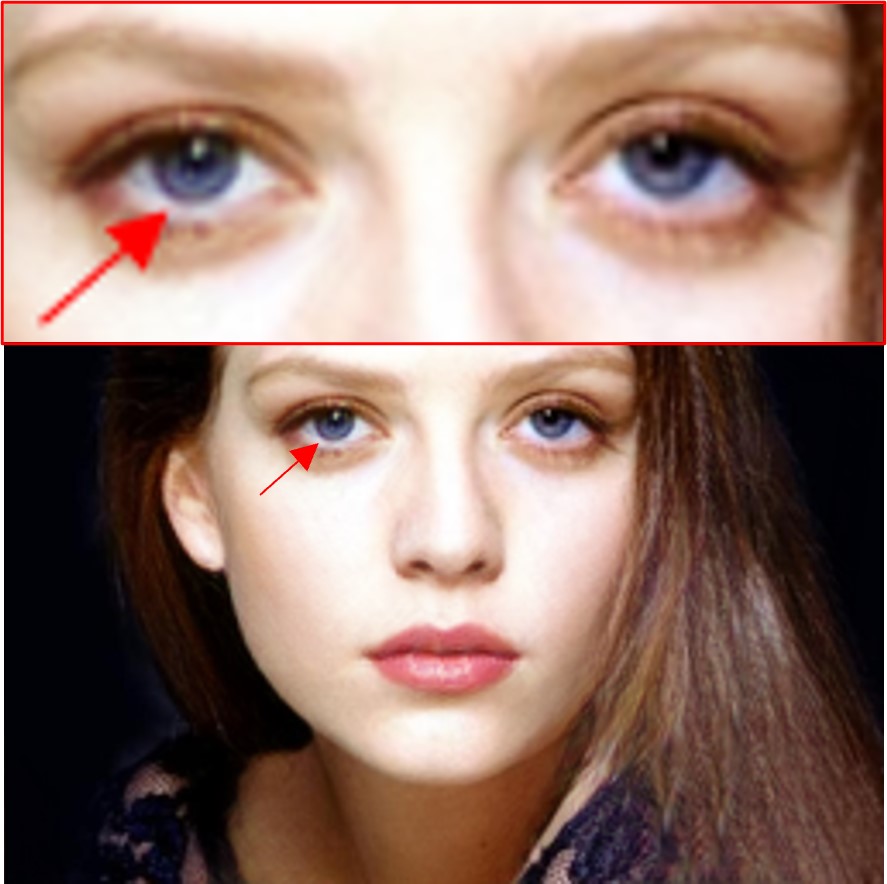}
    \includegraphics[width=0.1\textwidth]{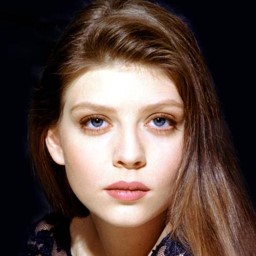}
    \\
    \makebox[0.1\textwidth]{\tiny Input}
    \makebox[0.1\textwidth]{\tiny WaveFill~\cite{yu2021wavefill}}
    \makebox[0.1\textwidth]{\tiny LAMA~\cite{wan2021high}}
    \makebox[0.1\textwidth]{\tiny MISF~\cite{li2022misf}}
    \makebox[0.1\textwidth]{\tiny MAT~\cite{li2022mat}}
    \makebox[0.1\textwidth]{\tiny RePaint~\cite{Lugmayr_2022_CVPR}}
    \makebox[0.1\textwidth]{\tiny CMT~\cite{ko2023continuously}}
    \makebox[0.1\textwidth]{\tiny Ours}
    \makebox[0.1\textwidth]{\tiny GT}
    \\
    \vspace{-0.7em}
    \caption{Comparisons with visualisations $(256 \times 256)$ showing that our results are more coherent in structure and sharper in texture and semantic details. \revise{The top three rows are from Places2~\cite{zhou2017places} and the bottom three rows are from CelebA-HQ~\cite{karras2017progressive}.}}
    \label{fig:comparison}
\end{figure*}

\subsection{Qualitative Comparison}
\label{exp:quali}

We showcase the qualitative image inpainting results on both datasets in Fig.~\ref{fig:comparison}. Each sample is the inpainted result where the mask ratio exceeds 40\%, to more intuitively demonstrate the advantages of SEM-Net in handling challenging cases. In facial inpainting, generating one eye in masked regions (masked eye) based on another eye in visible regions (visible eye) is more challenging than directly generating two eyes, because it requires the model to have a solid ability to capture long-range dependency to learn from another eye. Compared with current state-of-the-art techniques, SEM-Net successfully transfers features in the visible eye to the masked eye, including eyeball colour and shape, while preserving finer-grained features. In Places2, SEM-Net generates fewer artefacts and more coherent structures, such as the white lines in the road and the edges of coloured cardboard, ensuring the contextual consistency of the texture and structure of the image.

\begin{table}[htbp]
\captionsetup[table]{skip=4pt}
\vspace{-0.2cm}
\caption{Ablation studies of each component in $40\%-60\%$ mask ratio. Refer to supplementary material for all mask ratios.}
\vspace{-0.2cm}
	\centering
        \begin{adjustbox}{width=\linewidth}
		\begin{tabular}{@{\extracolsep{3pt}} c c c c c c | c c c c c@{}}
				\hline\hline
                    \multicolumn{1}{c}{\multirow{2}{*}{Net}} & 
				  \multicolumn{5}{c}{Components} & 
				\multicolumn{5}{c}{40\%-60\%}\T\B \\
					& MB & FN~\cite{zamir2022restormer} & SEFN & SBDM & PE & PSNR$\uparrow$ & SSIM$\uparrow$ & L1$\downarrow$ & FID$\downarrow$  & LPIPS$\downarrow$\T\B\\
				\hline\hline

        \textbf{\textcolor{black}{(a)}} & & & & & & 21.6134 & 0.7308 & 4.1254 & 8.1732 & 0.2464 \\
        \textbf{\textcolor{black}{(b)}} &\checkmark &\checkmark & & & & 21.7828 & 0.7587 & 3.9117 & 8.0742 & 0.2227\\
        \textbf{\textcolor{black}{(c)}} &\checkmark & &\checkmark & & & 22.0510 & 0.7682 & 3.7649 & 7.9871 & 0.2132 \\ 
        \textbf{\textcolor{black}{(d)}} &\checkmark &\checkmark & &\checkmark & & 21.9064 & 0.7653 & 3.7679 & 8.0214 & 0.2102 \\
        \textbf{\textcolor{black}{(e)}} &\checkmark & &\checkmark &\checkmark & & 22.0926 & 0.7692 & 3.7634  & 7.9174 & 0.2091  \\ 
        \textbf{\textcolor{black}{(f)}} &\checkmark &\checkmark & &\checkmark &\checkmark & 22.1776 & 0.7708 & 3.6747 & 7.9125 & 0.2095 \\ 
        \textbf{\textcolor{black}{(g)}} &\checkmark & &\checkmark &\checkmark &\checkmark & \textbf{22.1780} & \textbf{0.7725} & \textbf{3.6274} & \textbf{7.8915} & \textbf{0.2038}\B\\
        \hline
        \hline 
        \end{tabular}
        \end{adjustbox}
\label{tab:ablation_short}
\vspace{-0.3cm}
\end{table}

\begin{table}[htbp]
\captionsetup[table]{skip=4pt}
\caption{Comparison between our proposed SMB with transformer-based methods in $40\%-60\%$ mask ratio. Refer to supplementary material for all mask ratios.}
	\centering
            \scalebox{0.7}{
			\begin{tabu}{@{\extracolsep{3pt}} c  c c c c c c@{}}
				\hline\hline
				\multicolumn{1}{c}{\multirow{1}{*}{Input}} & \multicolumn{1}{c}{\multirow{2}{*}{Model}} & 
				\multicolumn{5}{c}{40\%-60\%}\T\B \\
				\cline{3-7} 
					Resolution& & PSNR$\uparrow$ & SSIM$\uparrow$ & L1$\downarrow$ & FID$\downarrow$ & LPIPS$\downarrow$ \T\B\\
				\hline\hline
        \multicolumn{1}{c}{\multirow{3}{*}{256*256}}& CSA~\cite{zamir2022restormer}   & 21.5362 & 0.7543 & 4.0471 & 8.1652 & 0.2326\\
        & SSA~\cite{dosovitskiy2020image} &\multicolumn{5}{c}{\multirow{1}{*}{Out of memory}}\\
         & SMB & \textbf{22.1776} & \textbf{0.7708} & \textbf{3.6747} & \textbf{7.9125} & \textbf{0.2095} \\  
        \hline
        \hline
        \multicolumn{1}{c}{\multirow{2}{*}{64*64}}& SSA~\cite{dosovitskiy2020image} & 20.1655 & 0.7265 & 5.2256 & 5.5547 & 0.1702\\
        & SMB & \textbf{20.1716} & \textbf{0.7352} & \textbf{5.1332} & \textbf{5.3158} & \textbf{0.1617}\B\\
        \hline
        \hline 
\end{tabu}}
\vspace{-0.6cm}
\label{tab:vsSAshort}
\end{table}

\begin{figure*}[t] \centering
    \includegraphics[width=0.10\textwidth]{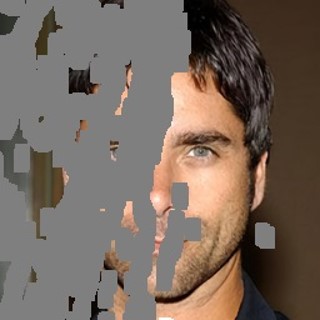}
    \includegraphics[width=0.10\textwidth]{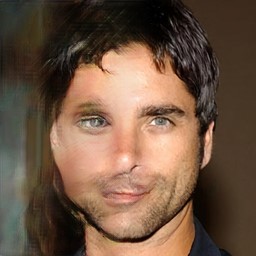}
    \includegraphics[width=0.10\textwidth]{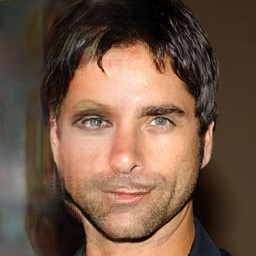}
    \includegraphics[width=0.10\textwidth]{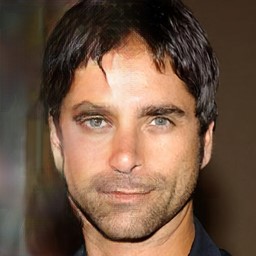}
    \includegraphics[width=0.10\textwidth]{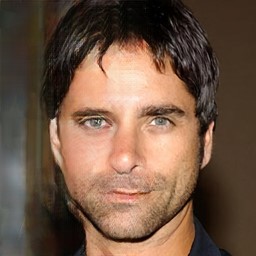}
    \includegraphics[width=0.10\textwidth]{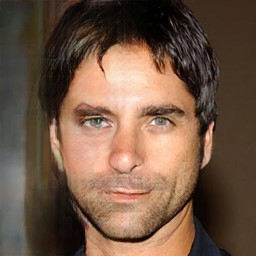}
    \includegraphics[width=0.10\textwidth]{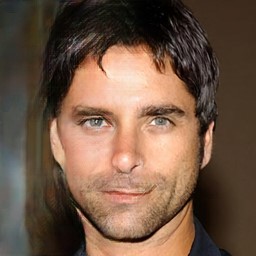}
    \includegraphics[width=0.10\textwidth]{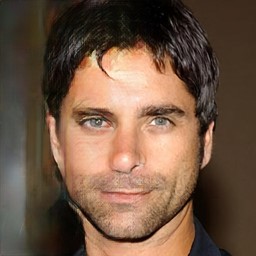}
    \includegraphics[width=0.10\textwidth]{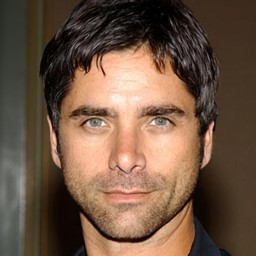}
    \\
    \makebox[0.10\textwidth]{\footnotesize Input}
    \makebox[0.10\textwidth]{\footnotesize a}
    \makebox[0.10\textwidth]{\footnotesize b}
    \makebox[0.10\textwidth]{\footnotesize c}
    \makebox[0.10\textwidth]{\footnotesize d}
    \makebox[0.10\textwidth]{\footnotesize e}
    \makebox[0.10\textwidth]{\footnotesize f}
    \makebox[0.10\textwidth]{\footnotesize g}   
    \makebox[0.10\textwidth]{\footnotesize GT}
    \vspace{-0.3cm}
    \caption{The qualitative visualisation of ablation studies. Zoom in for the details.}
    \vspace{-0.3cm}
    \label{fig:ablation}
\end{figure*}
\begin{figure*}[t] \centering
    \begin{tabular}{cl} 
        \rotatebox{90}{{\fontfamily{ptm}\selectfont $\quad$2560$\times$1920}} &     
        \includegraphics[width=0.23\textwidth]{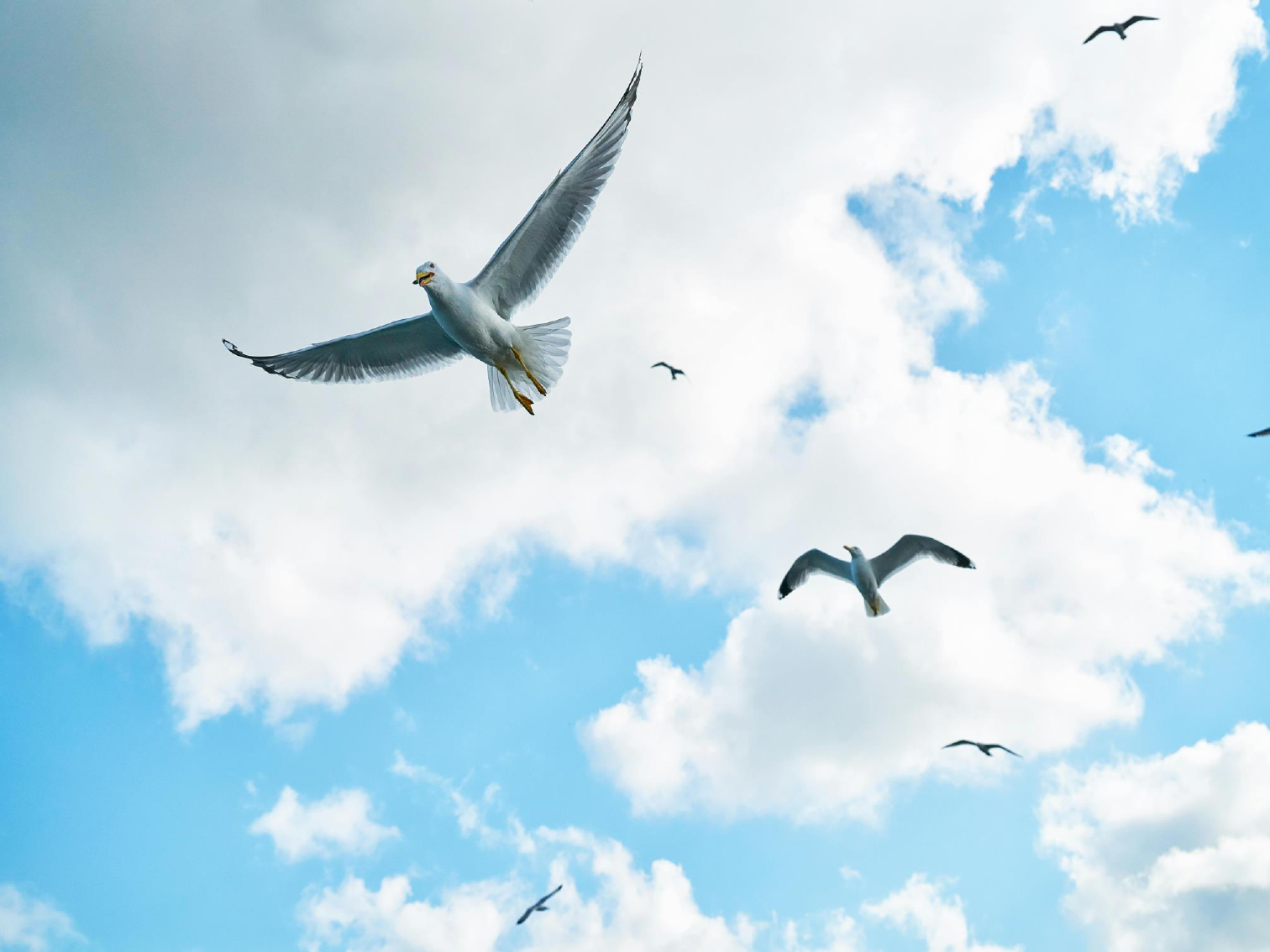}
        \includegraphics[width=0.23\textwidth]{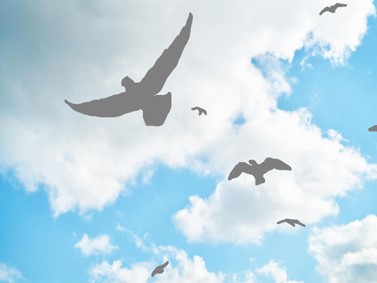}
        \includegraphics[width=0.23\textwidth]{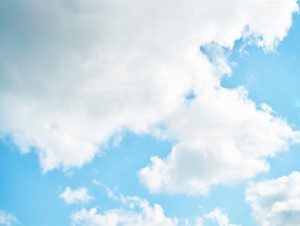}
        \includegraphics[width=0.23\textwidth]{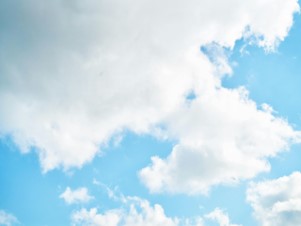}
        \\
    \end{tabular}
    \vspace{-0.2em}
    \\
    \begin{tabular}{cl} 
        \rotatebox{90}{{\fontfamily{ptm}\selectfont $\quad$2560$\times$1920}} &  
        \includegraphics[width=0.23\textwidth]{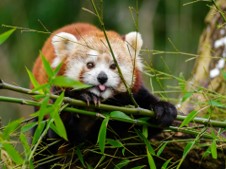}
        \includegraphics[width=0.23\textwidth]{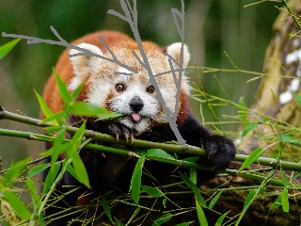}
        \includegraphics[width=0.23\textwidth]{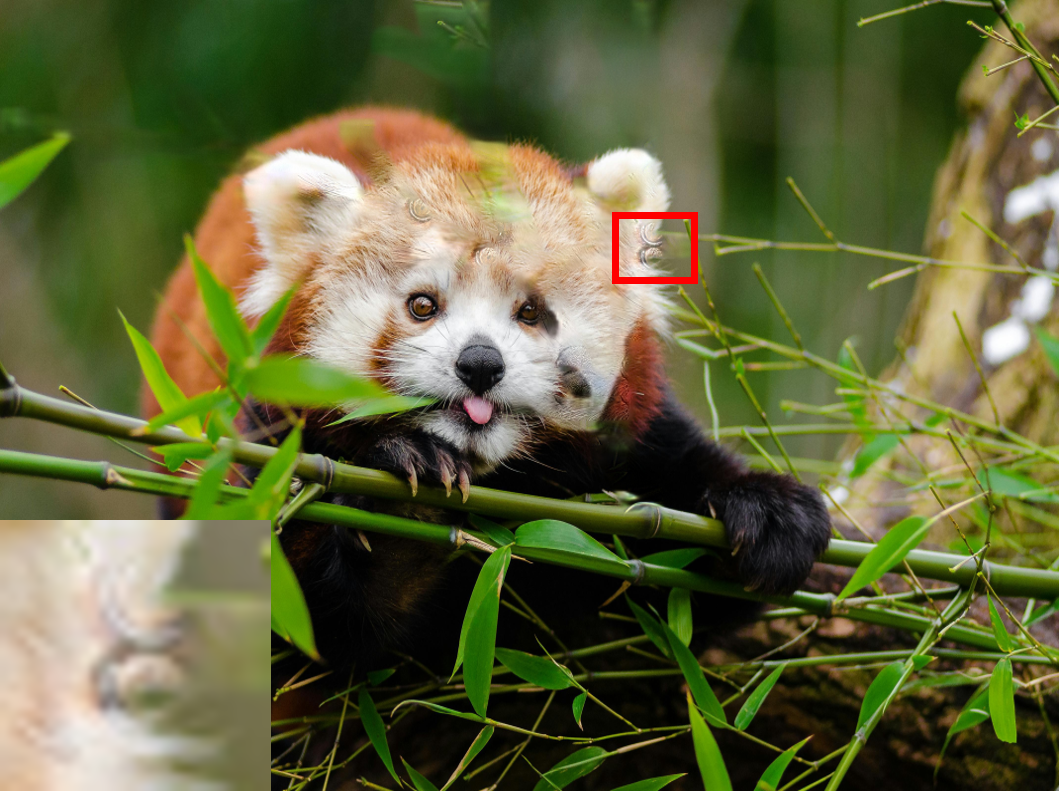}      
        \includegraphics[width=0.23\textwidth]{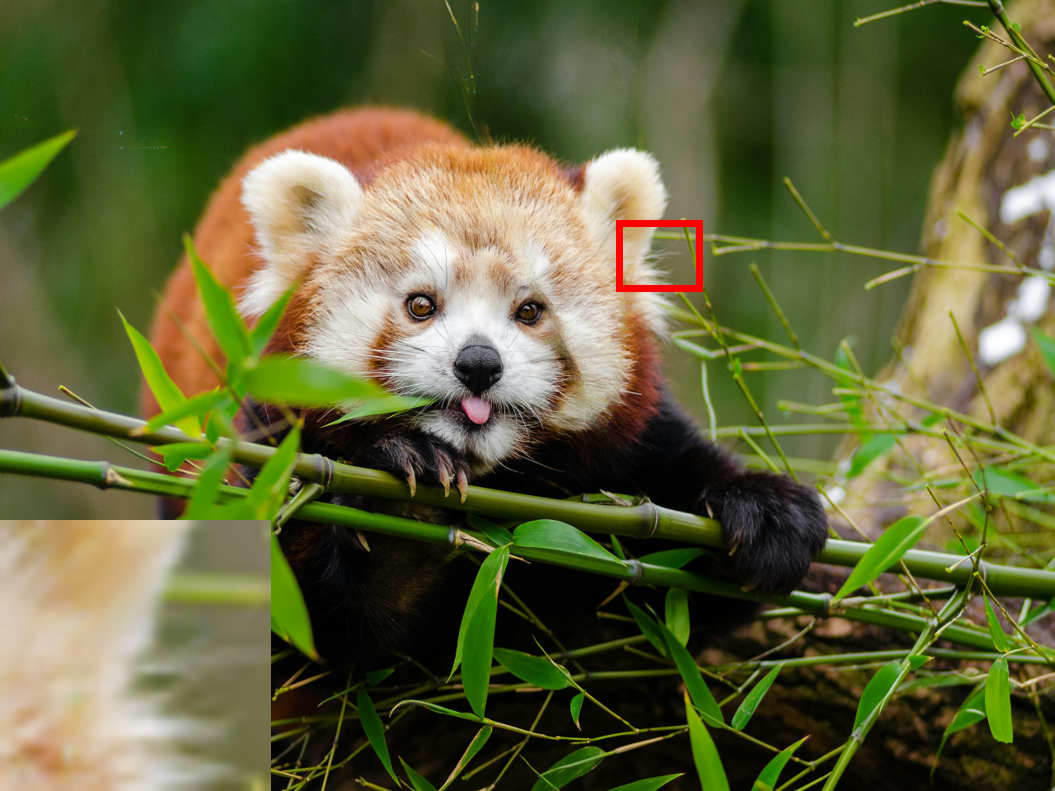} \\
    \end{tabular}
    \\
    \makebox[0.23\textwidth]{GT}
    \makebox[0.23\textwidth]{Masked Input}
    \makebox[0.23\textwidth]{HiFill}
    \makebox[0.23\textwidth]{Ours}
    \\
    \vspace{-0.2cm}
    \caption{Examples of generalisation to real-world high-resolution images of $2560 \times 1920$.} 
    \vspace{-0.3cm}
    \label{fig:large}
\end{figure*}

\vspace{-0.35cm}
\subsection{Ablation Study and Component Analysis}
\label{exp:abla}
\vspace{-0.1cm}
To efficiently verify the proposed modules, we followed~\cite{cui2022selective} to conduct the ablation and component analysis experiments on a lighter version of SEM-Net with the halved number of SMB in each U-Net stage. Each experiment is trained on CelebA-HQ for 30K iterations.

\vspace{0.15cm}
\noindent \textbf{Improvement in Each Component.} Tab.~\ref{tab:ablation_short} and Fig.~\ref{fig:ablation} present the improvement of each component quantitatively and qualitatively. Based on the U-Net shape baseline (Tab.~\ref{tab:ablation_short}{\textcolor{red}{a}}), integrating the Mamba Block (MB) and Feedforward Network (FN)~\cite{zamir2022restormer} (Tab.~\ref{tab:ablation_short}{\textcolor{red}{b}}) results in noticeable improvements across all metrics.
Fig.~\ref{fig:ablation}{\textcolor{red}{d$\rightarrow$b}} and Fig.~\ref{fig:ablation}{\textcolor{red}{e$\rightarrow$c}} shows that degrading SBDM, model struggle in capture the relations of vertically adjacent pixels, resulting in artefacts between the left eyebrow and left eye.
Fig.~\ref{fig:ablation}{\textcolor{red}{b$\rightarrow$c}}, Fig.~\ref{fig:ablation}{\textcolor{red}{d$\rightarrow$e}} and Fig.~\ref{fig:ablation}{\textcolor{red}{f$\rightarrow$g}} revealed the effect of SEFN by resulting sharper jaw and less artefacts, demonstrated by the improvement in SSIM score. 
Tab.~\ref{tab:ablation_short}{\textcolor{red}{d$\rightarrow$f}} and Tab.~\ref{tab:ablation_short}{\textcolor{red}{e$\rightarrow$g}} showcase that introducing positional embedding significantly improves L1 and PSNR in larger masks, which is evidenced by the clearer texture at the mouth and eye. 

\vspace{-0.3cm}
\subsection{Comparing SMB with Transformer Blocks.}
We evaluate the effectiveness of our proposed SMB in image representation learning by comparing it with two typical and widely used transformer blocks that claimed to have strong capability in capturing LRDs: channel-wise self-attention~\cite{zamir2022restormer} and Spatial-wise self-attention (SSA) \cite{dosovitskiy2020image}. For fair comparisons, all models use vanilla feedforward networks~\cite{zamir2022restormer} instead of our novel SEFN, with only differences between SMB, CSA and SSA. From Table.~\ref{tab:vsSAshort}, we observe that our SMB consistently outperforms two distinct transformer blocks across all metrics in all mask ratios. In addition, our SMB is shown to be efficient enough to process original resolution ($256\times256$) images while SSA can only be trained on the degraded $64\times64$ images with a single A100 due to its significant computational cost. Furthermore, compared with the diffusion-based models~\cite{Rombach_2022_CVPR,Lugmayr_2022_CVPR} with a very long inference time~\cite{chang23design}, our model has better performance while the inference time is still in milliseconds, which is suitable for real-time scenarios (shown in Tab.~\ref{tab:long}). 

\subsection{Generalisation Ability}
\label{exp:gen}

\vspace{0.15cm}
\noindent \textbf{Unseen High Resolution Images.}
We examine the scalability and generalizability of SEM-Net trained on $256\times256$ Places2 images in processing unseen images of higher resolution, since these abilities are crucial for practical applications where image resolutions can significantly vary. Fig.~\ref{fig:large} showcases examples of unseen real-world high-resolution applications. \revise{While \cite{yi2020contextual} performs similarly with larger masks, its upsampling strategy causes narrow mask drifting, leading to artefacts. SEM-Net, by modelling at the pixel level, captures finer details without artefacts, offering the community a better, more resource-efficient solution for processing large-resolution images.} More examples with different resolutions are included in the supplementary.

\label{exp:delur}
\begin{table}[t]
\begin{center}
\caption{\small Performance in generalising to image motion deblurring task. Our SEM-Net is trained only on the GoPro dataset~\cite{gopro2017} and directly applied to the HIDE~\cite{shen2019human}.}
\label{tab:deblurring}
\setlength{\tabcolsep}{3mm}
\scalebox{0.69}{
\begin{tabular}{l c c | c c }
\toprule[0.15em]
 & \multicolumn{2}{c}{\textbf{GoPro}~\cite{gopro2017}} & \multicolumn{2}{c}{\textbf{HIDE}~\cite{shen2019human}} \\
 \textbf{Method} & PSNR $\uparrow$ & SSIM $\uparrow$ & PSNR $\uparrow$ & SSIM $\uparrow$ \\
\midrule[0.15em]
\small{DeblurGAN-v2 \cite{deblurganv2}}    & 29.55 & 0.934 & 26.61 & 0.875 \\
Shen \etal \cite{shen2019human}     & - & -             & 28.89 & {0.930}  \\
Gao \etal \cite{gao2019dynamic}     & 30.90 & {0.935} &  29.11 & {0.913}  \\
DBGAN \cite{zhang2020dbgan}         & 31.10 & {0.942} & 28.94 & {0.915}   \\
MT-RNN \cite{mtrnn2020}             & 31.15 & {0.945} & 29.15 & {0.918}  \\
DMPHN \cite{dmphn2019}              & 31.20 & {0.940}  & 29.09 & {0.924} \\
Suin \etal \cite{Maitreya2020}      & 31.85 & {0.948} & 29.98 & {0.930} \\
SPAIR~\cite{purohit2021spatially_spair} & 32.06 & {0.953} & 30.29 & {0.931} \\
MIMO-UNet+~\cite{cho2021rethinking_mimo} & 32.45 & {0.957} & 29.99 & {0.930}\\
IPT~\cite{chen2021IPT} & 32.52 & {-} & - & - \\
MPRNet~\cite{Zamir_2021_CVPR_mprnet} & {32.66} & {{0.959}} & {30.96} & {{0.939}} \\
HINet~\cite{chen2021hinet} & {32.71} & {{0.959}} & {30.32} & {{0.932}} \\
Restormer~\cite{zamir2022restormer} & 32.92 & \underline{0.961} &\textbf{31.22} &{\textbf{0.942}}\\
Stripformer~\cite{tsai2022stripformer} & \underline{33.08} & \textbf{0.962} &  31.03 & 0.940\\

\bottomrule[0.1em]
Ours &  \textbf{33.11} & \textbf{0.962} & \underline{31.12} & \underline{0.941} \\
\bottomrule[0.1em]
\end{tabular}}
\vspace{-0.6cm}
\label{tab:deblur}
\end{center}
\end{table}

\vspace{-0.15cm}
\noindent \textbf{Low-level Vision Tasks.}
To further evaluate the capability of representation learning and generalisation ability of SEN-Net, we directly apply SEN-Net to another low-level vision task, image motion deblurring, through the necessary learning of the residual between clear images and blurred images without any other task-specific modifications. Tab.~\ref{tab:deblur} shows that SEM-Net overall outperforms the restoration models on two synthetic benchmark datasets GoPro~\cite{gopro2017} and HIDE~\cite{shen2019human}. Especially on GoPro, SEM-Net improves PSNR by 0.19  compared to the strong restoration baseline model Restormer~\cite{zamir2022restormer}. Notably, our SEM is trained on GoPro and directly applied to HIDE, without progressive learning~\cite{zamir2022restormer} or Test-time Local Converter~\cite{chu2022improving} such external optimization, showcasing strong generalization ability. Refer to supplementary materials for qualitative results.

\section{Conclusion and Discussion}
We present an SSM-based image inpainting model, SEM-Net, which demonstrates strong capabilities in capturing LRDs and addresses the challenge of lack of spatial awareness in SSMs. We propose two key designs, SMB and SEFN, for improved image representation learning. 
\revise{With these designs, our model outperforms state-of-the-art approaches on two image inpainting datasets, especially on CelebA-HQ. This could be due to dataset characteristics, CelebA-HQ’s structured, human-centric images benefit more from our model’s ability to capture long-range dependencies and spatial awareness, shown in quantitative results. Also, we showcases strong generalizability to higher-resolution images and another low-level visual task, image deblurring. 
Our future work aims to build a controllable image inpainting model based on the proposed SMB to handle large-resolution images.}


{\small
\bibliographystyle{ieee_fullname}
\bibliography{egbib}
}

\end{document}